\documentclass[lettersize,journal]{IEEEtran}
\usepackage{amsmath,amsfonts}
\usepackage{algorithmic}
\usepackage{algorithm}
\usepackage{array}
\usepackage{textcomp}
\usepackage{stfloats}
\usepackage{url}
\usepackage{verbatim}
\usepackage{graphicx}
\usepackage{hyperref}
\usepackage[table,xcdraw]{xcolor} 
\usepackage[style=ieee]{biblatex}
\usepackage{csquotes}
\usepackage{soul}
\usepackage{enumerate}

\usepackage{booktabs}
\usepackage{multirow}
\usepackage[switch]{lineno}
\usepackage{xr}

\usepackage{titling}
\newcommand{\beginsupplement}{%
        \setcounter{table}{0}
        \renewcommand{\thetable}{S\arabic{table}}%
        \setcounter{figure}{0}
        \renewcommand{\thefigure}{S\arabic{figure}}%
        \setcounter{section}{0}
        \renewcommand{\thesection}{S\arabic{section}}%
     }

\begin{document} \label{main}

\title{Tuned Compositional Feature Replays for Efficient Stream Learning}

\date{}
\author {
    Morgan B. Talbot,\textsuperscript{\rm 1,2,3,4*}
    Rushikesh Zawar,\textsuperscript{\rm 5,2,6*}
    Rohil Badkundri,\textsuperscript{\rm 5,2,4} \\
    Mengmi Zhang,\textsuperscript{\rm 4,2,7,8,9\dag}
    and Gabriel Kreiman\textsuperscript{\rm 2,3,4\dag}\\
    \small * Equal contribution \quad $\dag$ Corresponding authors\\
    \small Address correspondence to mengmi.zhang@ntu.edu.sg; gabriel.kreiman@tch.harvard.edu \\
    \small \textsuperscript{\rm 1} Harvard-MIT Program in Health Sciences and Technology (HST), Cambridge, MA, USA \\
    \small \textsuperscript{\rm 2} Boston Children's Hospital, Boston, MA, USA \\
    \small \textsuperscript{\rm 3} Harvard Medical School, Boston, MA, USA \\
    \small \textsuperscript{\rm 4} Center for Brains, Minds, and Machines (CBMM), Cambridge, MA, USA \\
    \small \textsuperscript{\rm 5} Harvard University, Cambridge, MA, USA \\ 
    \small \textsuperscript{\rm 6} Birla Institute of Technology and Science at Pilani, Pilani, India \\
    \small \textsuperscript{\rm 7} Agency for Science, Technology, and Research (A*STAR) Center for Frontier AI Research (CFAR), Singapore \\
    \small \textsuperscript{\rm 8} A*STAR Center for Infocomm Research (I2R), Singapore \\
    \small \textsuperscript{\rm 9} School of Computer Science and Engineering, Nanyang Technological University, Singapore \\
}




\maketitle

\begin{abstract}
Our brains extract durable, generalizable knowledge from transient experiences of the world. Artificial neural networks come nowhere close to this ability. When tasked with learning to classify objects by training on non-repeating video frames in temporal order (online stream learning), models that learn well from shuffled datasets catastrophically forget old knowledge upon learning new stimuli. We propose a new continual learning algorithm, Compositional Replay Using Memory Blocks (CRUMB), which mitigates forgetting by replaying feature maps reconstructed by combining generic parts. CRUMB concatenates trainable and re-usable \enquote{memory block} vectors to compositionally reconstruct feature map tensors in convolutional neural networks. Storing the indices of memory blocks used to reconstruct new stimuli enables memories of the stimuli to be replayed during later tasks. This reconstruction mechanism also primes the neural network to minimize catastrophic forgetting by biasing it towards attending to information about object shapes more than information about image textures, and stabilizes the network during stream learning by providing a shared feature-level basis for all training examples. These properties allow CRUMB to outperform an otherwise identical algorithm that stores and replays raw images, while occupying only 3.6\% as much memory. We stress-tested CRUMB alongside 13 competing methods on 7 challenging datasets. To address the limited number of existing online stream learning datasets, we introduce 2 new benchmarks by adapting existing datasets for stream learning. With only 3.7-4.1\% as much memory and 15-43\% as much runtime, CRUMB mitigates catastrophic forgetting more effectively than the state-of-the-art. Our code is available at  \href{https://github.com/MorganBDT/crumb.git}{https://github.com/MorganBDT/crumb.git}.
\end{abstract}


\section{Introduction}
\label{sec:intro}

\begin{figure}[t]
    \begin{center}
        \includegraphics[width=8cm, height = 5cm]{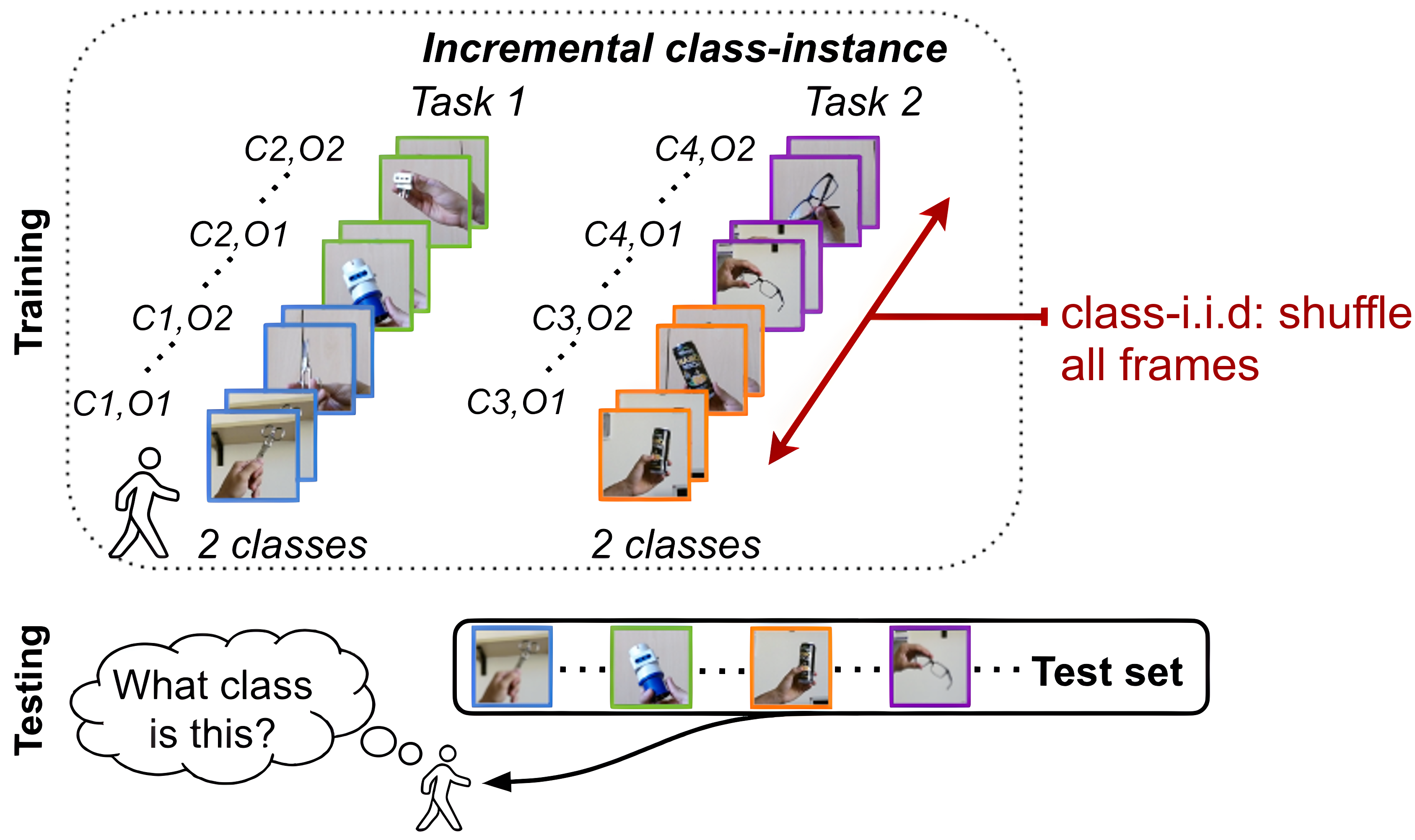}
    \end{center}
    \vspace{-4mm}
    \caption{\textbf{Schematic of online stream learning protocols.} 
    For each task, the model learns to classify a set of new classes (C1, C2, etc. in figure) while training on video clips of several objects from each class (O1, O2) for only one epoch. During testing, the model has to classify images from all seen classes without knowing task identity. In the class-instance training protocol, the order of video clips is shuffled but the order of frame images is preserved within each clip. In the class-i.i.d. training protocol, all images within each task are randomly shuffled. Class-i.i.d. is the only option for datasets such as ImageNet that consist of standalone images and not video clips.  
    }\label{fig:protocols}\vspace{-4mm}
\end{figure}

\IEEEPARstart{H}{umans} adapt to new and changing  environments by learning rapidly and continuously. Previously learned skills and experiences are retained even as they are transferred and applied to new tasks, which are learned from a stream of highly temporally correlated stimuli and without direct access to past experiences. In contrast, in standard class-incremental image classification settings in continual learning, neural networks are presented with images that are independently and identically distributed (i.i.d.), with multiple presentations of each image ~\cite{mccloskey1989catastrophic,ratcliff1990connectionist,french1999catastrophic}. To better emulate a human learning environment, or that of an autonomous robot that must learn in real time, we focus on a challenging and 
realistic variant of class-incremental learning ---
\emph{online stream learning}. Online stream learning has two key characteristics (Fig.~\ref{fig:protocols}): (a) the input is in the form of video streams with highly temporally correlated frames, and (b) each training example is presented only once: no repeated presentations of old data are allowed.

In online stream learning settings, current machine learning systems tend to fail to retain good performance on previously learned tasks, exhibiting catastrophic forgetting ~\cite{hayes2020remind,kemker2018measuring,maltoni2019continuous}. Catastrophic forgetting is a pervasive problem in continual learning settings for both deep neural networks \cite{de2021continualsurvey} and other models such as linear regression \cite{evron2022catastrophic} and self-organizing maps \cite{vaidya2022reducing}, and can also cause neural models to be biased towards more recently encountered training data \cite{shao2022overcoming}. One strategy for overcoming catastrophic forgetting is to store a copy of all or most encountered training examples for later replay, effectively converting to an offline learning paradigm \cite{prabhu2020gdumb}. This approach, however, often requires an impractically large amount of memory \cite{kirkpatrick2017overcoming}. Moreover, much of the information in raw images is redundant, with many pixel values needed to represent each feature-level concept relevant to classification. Finally, storing old training data might also be undesirable from a data security or privacy standpoint, such as in hospitals and other healthcare settings \cite{lee2020clinical}.  

To address both memory inefficiency and data privacy concerns while achieving state-of-the-art online stream learning performance, we propose a new continual learning approach, Compositional Replay Using Memory Blocks (CRUMB) (Fig.~\ref{fig:our_model}). In our method, each new image is processed by the early layers of a convolutional neural network (CNN) to produce a feature map tensor. The feature map is decomposed by slicing it into chunks, each of which is a vector of feature activations at a specific spatial location. Each chunk is then replaced by the most cosine-similar row (\enquote{memory block}) of a trainable ``codebook matrix.'' This mechanism encodes images as a composition of discrete feature-level concepts, some of which appear to have semantic interpretations. Storage of a complete training example for replay requires keeping only the indices of the memory blocks needed to reconstruct the original feature map, along with the class label, occupying only 3.6\% of the memory footprint of a raw image. During replay, feature maps reconstructed via stored indices are fed to the later layers of the CNN, such that these layers are trained on both stored and newly encountered images to learn new tasks while retaining previous knowledge. 

Our key contributions are:
\begin{itemize}
  \item \textbf{Trainable Compositional Replay.} We propose a new compositional feature-level replay algorithm, CRUMB, for online stream learning. The composition mechanism is end-to-end trainable and reusable. CRUMB's codebook of memory blocks captures the essential components needed for reconstructing feature maps. During the pretraining phase, the memory block mechanism primes the CNN for stream learning with high accuracy and induces a beneficial bias towards object shapes. Using memory blocks as a shared basis for new and recalled examples helps stabilize the network during stream learning. 
  \item \textbf{Reduced forgetting.} We tested CRUMB on 7 continual learning datasets alongside 13 competing methods, showing that CRUMB typically outperforms state-of-the-art approaches by large margins. 
  \item \textbf{Superior Efficiency.} Storing $n$ compositional feature maps for replay prevents catastrophic forgetting substantially more effectively than storing $n$ raw images, while only requiring about 3.6\% as much memory. Additionally, compared with the next most accurate method (REMIND \cite{hayes2020remind}), CRUMB requires only about 15-43\% as much training runtime, and occupies only 3.7-4.1\% of REMIND's peak memory footprint.
  \item \textbf{New Benchmarks.} We adapted 2 datasets, Toybox \cite{wang2018toybox} and iLab \cite{borji2016ilab}, to introduce new online stream learning benchmarks. All benchmark details along with source code, results, and data are available at \href{https://github.com/MorganBDT/crumb.git}{https://github.com/MorganBDT/crumb.git}.
\end{itemize}

\section{Related work}

\subsection{Weight regularization}
Weight regularization methods typically store weights trained on previous tasks and impose constraints on subsequent weight updates to minimize catastrophic forgetting \cite{li2017learning,chaudhry2018efficient,he2018overcoming,kirkpatrick2017overcoming,zenke2017continual,lee2017overcoming, kong2023overcoming}. However, storing the importance of the millions of parameters required by state-of-the-art recognition models across all previous tasks is costly ~\cite{kirkpatrick2017overcoming,wen2018few}.
Moreover, empirical comparisons suggest that weight regularization methods typically do not mitigate catastrophic forgetting as effectively as architecture adaptation and replay methods \cite{mai2022online}.

\subsection{Architecture adaptation}
Architecture adaptation methods expand or re-organize the structure of their neural networks to accommodate new tasks to be learned. Approaches include adding groups of new neurons (which does not always scale well) \cite{li2017learning,he2018overcoming,kirkpatrick2017overcoming,zenke2017continual,lee2017overcoming, zhang2022self}, isolating parts of a larger neural network for each task \cite{fernando2017pathnet, rajasegaran2019random, serra2018overcoming, adel2019continual, bricken2023sparse}, compressing parameters in a consolidation phase \cite{schwarz2018progress}, and pruning neurons or weights for later re-use. Pruning approaches include L1 regularization and activity threshold-based sparsification of neurons \cite{golkar2019continual}, and combining pruning of weights with parameter importance-based regularization \cite{peng2021overcoming}. Neuron pruning/re-use can also be combined with the addition of new neurons to improve performance and enable increased flexibility \cite{gao2022efficient}. All of these approaches add significant complexity, and some require explicit labelling of task identities, which is not feasible in many online learning applications. 

\subsection{Image and feature replay}
In replay methods, images or features from previous tasks are stored and later retrieved or re-generated to be shown to the model to prevent forgetting \cite{wu2019large,rebuffi2017icarl,aljundi2019gradient, chaudhry2018efficient, nguyen2017variational,lopez2017gradient,bang2021rainbow}. Replay can be highly effective, but comes with some caveats. Relying on replaying limited sets of stored examples can lead to overfitting. Storing a large number of raw images for replay is memory-intensive. To limit memory requirements, generative replay systems combine data from new tasks with synthetic data produced by generative models to resemble previously encountered stimuli
~\cite{shin2017continual,robins1995catastrophic,liu2020mnemonics,atkinson2018pseudo,liu2020generative,shen2021generative,van2021class}. However, the generative models needed to create adequate synthetic data remain large, memory-intensive, and difficult to train \cite{wen2018few}.

Other replay methods save memory by storing raw or compressed feature maps from intermediate CNN layers~\cite{pellegrini2020latent, hayes2020remind}, or generate synthetic examples by sampling from simple feature-level probability distributions for each class \cite{zhang2021memory}. REMIND ~\cite{hayes2020remind} achieves high performance in online stream learning by compressing feature maps using a product quantizer \cite{jegou2010product}. However, the product quantizer is trained by performing k-means clustering on a large subset of training data stored in memory, a process that scales poorly for increasingly large datasets. In contrast, CRUMB's differentiable codebook is trained by backpropagation alongside other network parameters, dramatically reducing memory requirements for codebook initialization. 

CRUMB's feature-based replay mechanism is inspired by biological replay observed in the hippocampus and other brain areas \cite{o2010play,lewis2011overlapping,eichenlaub2020replay}, and by complementary learning systems theory \cite{mcclelland1995there}. Recent work has explored modeling the hippocampus and neocortex as separate neural networks that interact via distillation losses and other mechanisms \cite{peng2023lifelong}. In contrast to storing knowledge implicitly in a short-term memory network, CRUMB's memory blocks and replay buffer store short-term memories that represent individual training examples, and interact with the CNN via replay to facilitate longer-term memory storage and consolidation. 

\section{Methods}

\subsection{Online stream learning benchmarks}

\subsubsection{Training protocols} \label{sec:protocols} 
We consider two online class-incremental settings: class-instance and class-i.i.d. \cite{hayes2020remind}
(Fig.~\ref{fig:protocols}). 

\textbf{Class-instance}. Each task contains short video clips of different objects from two or more classes, and the video clips are presented one after another in random order within each task without repetition. An ideal learning algorithm in this setting would be stable enough to remember prior tasks while being sufficiently plastic to learn generalizable class boundaries for new classification tasks, despite encountering many highly correlated images of each object before moving on to the next. 

\textbf{Class-i.i.d.}. 
Images/video frames are randomly shuffled within each task but not interspersed among tasks, and are shown only once like in class-instance. This is a less challenging protocol, and should not be considered stream learning in the strictest sense because the shuffling of images in each task destroys any temporal structure among images. 

In both settings, our model and all competing baseline models are allowed to train for many epochs on the first task, but are restricted to viewing each image from subsequent tasks only once. This emulates real-time acquisition of training data that cannot be stored except in a limited-capacity replay buffer. 

\subsubsection{Stream learning benchmark datasets}\label{protocoldata}
We evaluated our model on five video datasets (class-instance and class-i.i.d. protocols), and two image datasets (class-i.i.d. only). For all datasets, we used different task and example orderings across training runs. A global holdout test set of images/frame sequences was used for all runs. To help address the limited number of online stream learning benchmarks, we adapted two datasets designed for studying object transformations, Toybox~\cite{wang2018toybox} and iLab~\cite{borji2016ilab}, for online stream learning. 

The \textbf{CORe50 video dataset~\cite{lomonaco2017core50}} contains images of 50 objects in 10 classes. Each object has 11 instances, which are 15 second video clips of the object under particular conditions and poses. We followed ~\cite{hayes2020remind} for the training and testing data split, and sampled each video at 1 frame per second (FPS). 
 
The \textbf{Toybox video dataset~\cite{wang2018toybox}} 
contains videos of toy objects from 12 classes. We used a subset of the  dataset containing 348 toy objects, each of which has 10 instances containing different patterns of object motion. We sampled each instance at 1 FPS, resulting in 15 images per instance per object. We chose 3 of the 10 instances for our test set, leaving 7 instances for training. 

The \textbf{iLab (iLab-2M-Light) video dataset~\cite{borji2016ilab}}
contains videos of toy vehicles from 14 classses. We used a subset of the dataset containing 392 vehicles, with 8 instances (backgrounds) per object and 15 images per instance. We chose 2 of the 8 instances for our test set. 

The \textbf{iCub (iCubWorld Transformations) video dataset~\cite{pasquale2016object}} contains videos taken by the iCub robot of 20 classes of household objects undergoing viewpoint transformations. We used isolated rotation, scaling, and background transformations as our training set, and the provided ``MIX'' sequence (which combines all transformations) as our test set.

The \textbf{iLab+CORe50 video dataset} combines iLab and CORe50 to create a stream learning benchmark with 24 distinct classes. All iLab classes are learned before CORe50, introducing a mild domain shift. We uniformly subsample iLab to balance the number of images per class with CORe50.

To evaluate our model in long-range online class-incremental learning with many more classes than the video datasets described above, we also include results on two image datasets. The standard \textbf{Online-CIFAR100 image dataset~\cite{krizhevsky2009learning}} is split into 20 tasks with 5 classes each, while the standard \textbf{Online-Imagenet image dataset~\cite{deng2009imagenet}} is split into 10 tasks with 100 classes each. Class-instance training is not applicable to image datasets because they do not consist of videos. 

\subsubsection{Baseline algorithms for comparison} \label{sec:baselines}
All baseline algorithms use the same training protocols as CRUMB. CRUMB and most baselines use a SqueezeNet CNN pretrained on ImageNet \cite{iandola2016squeezenet}, but due to implementation constraints AAN\cite{liu2021adaptive}, CoPE\cite{de2020continual}, GSS\cite{aljundi2019gradient}, LwF\cite{li2017learning}, RM\cite{bang2021rainbow}, and Stable SGD\cite{mirzadeh2020understanding} use non-pretrained ResNet models \cite{he2016deep}. We re-implemented some methods due to varying code availability. CRUMB and all baselines are implemented using the PyTorch library \cite{paszke2019pytorch}. 

\noindent \textbf{Weight Regularization:} We compared against Elastic Weight Consolidation (EWC) \cite{kirkpatrick2017overcoming}, Synaptic Intelligence (SI) \cite{zenke2017continual}, Memory Aware Synapses (MAS) \cite{aljundi2018memory}, Learning without Forgetting (LwF) \cite{li2017learning}, and Stable SGD \cite{mirzadeh2020understanding}. 

\vspace{0.5mm}
\noindent \textbf{Memory Distillation and Replay}: 
We compared against Gradient Episodic Memory (GEM) \cite{lopez2017gradient}, Incremental Classifier and Representation Learner (iCARL)~\cite{rebuffi2017icarl}, Bias Correction (BiC) \cite{wu2019large}, Gradient Sample Selection (GSS) \cite{aljundi2019gradient}, Continual Prototype Evolution (CoPE) \cite{de2020continual}, Adaptive Aggregation Network (AAN) \cite{liu2021adaptive}, REMIND \cite{hayes2020remind}, and Rainbow Memory (RM) \cite{bang2021rainbow}.

\noindent The \textbf{Lower bound} is trained sequentially over all tasks without any measures to avoid catastrophic forgetting.

\noindent The \textbf{Upper bound} is trained offline on shuffled images from both the current and all previous tasks over multiple epochs. 

\noindent \textbf{Chance} predicts class labels by randomly choosing 1 out of the total of $C_t$ classes seen in or before current task $t$. 

\vspace{-2mm}
\subsection{Proposed algorithm: CRUMB}\label{sec:modeloverview}

\begin{figure*}[ht]
    \begin{center}
        \includegraphics[width=16cm]{MMfigure/Fig2_highres.png}
    \end{center}
    \vspace{-5mm}
    \caption{\textbf{Schematic illustration of CRUMB, the proposed algorithm for online stream learning.} The model consists of a CNN ($\mathbf{F(\cdot)}$ for early layers and $\mathbf{P(\cdot)}$ for later layers) and a codebook matrix $\mathbf{B}$ used for compositional reconstruction of feature-level activation tensors (feature maps $\mathbf{Z}$). Each row $\mathbf{B_k}$ of $\mathbf{B}$ is a ``memory block'' vector. CRUMB uses the feature extractor $\mathbf{F(\cdot)}$ to produce an initial feature map, then determines which memory blocks to retrieve from $\mathbf{B}$ based on a cosine-similarity addressing mechanism. The feature maps reconstructed from the memory blocks ($\mathbf{\widetilde{Z}}$), and the original feature maps ($\mathbf{Z}$), are used to obtain separate classification losses from the same classifier network $\mathbf{P(\cdot)}$ (\enquote{\textbf{codebook-out loss}} and \enquote{\textbf{direct loss}}, respectively). Only codebook-out loss is used for weight updates during stream learning, although the two losses are added in a weighted sum to calculate the total loss during pretraining. To avoid catastrophic forgetting, we store the row indices of retrieved memory blocks along with class labels for example images from each task. In later tasks, following each batch of new images, we \enquote{replay} a batch of old feature maps to $\mathbf{P(\cdot)}$ after reconstructing them using stored memory block indices. 
    } \label{fig:our_model}
\end{figure*}

We propose a new continual learning algorithm, Compositional Replay Using Memory Blocks (CRUMB). CRUMB consists of a 2-dimensional CNN augmented by an $n \times d$ codebook matrix $B$. A schematic of CRUMB is shown in Fig.~\ref{fig:our_model}, with further details described in algorithm~\ref{algo:algorithm1}. CRUMB extracts a feature map from each given image using the early layers of a pre-trained CNN, and stores a subset of the feature maps in a buffer. When CRUMB later encounters a new task, it avoids catastrophic forgetting of previous tasks by replaying stored feature maps of images from those tasks to the later layers of the network. To reduce memory requirements, CRUMB uses its codebook matrix $B$ to reconstruct each feature map. Rows of $B$ (\enquote{memory blocks}) are concatenated to form tensors that approximate the original feature maps, and only the indices of activated memory blocks need to be stored to enable later reconstruction. All computations from memory block reconstruction forward are differentiable, allowing $B$ to be learned alongside the CNN weights. 

\subsubsection{Feature extraction and classification}\label{subsection:fec}

CRUMB's CNN backbone is split into two nested functions. The early layers of the network comprise $F(\cdot)$, a \enquote{feature extractor,} while the remaining, later layers comprise $P(\cdot)$, a classifier. Since early convolutional layers of CNNs are highly transferable \cite{yosinski2014transferable}, the parameters of $F(\cdot)$ are pretrained for image classification using ImageNet~\cite{deng2009imagenet} and then fixed during stream learning. CRUMB passes each training image through feature extractor $F(\cdot)$ to obtain feature map $Z$, of size $s \times w \times h$ (number of features, width, height). $Z$ is reconstructed using $B$ to form $\widetilde{Z}$, and a class prediction output can then obtained as $P(\widetilde{Z})$. The parameters of $P(\cdot)$ are initially pretrained alongside $F(\cdot)$ on ImageNet using standard methods, but also undergo additional ImageNet pretraining with an objective incorporating predictions on both $Z$ and $\widetilde{Z}$. Prior to stream learning, only the final layer of $P(\cdot)$ is randomly reinitialized to reflect the number of classes to be learned during streaming.
\subsubsection{Reconstructing feature maps from memory}\label{subsection:reconstructmem}

CRUMB produces reconstructed feature map $\widetilde{Z}$ using only $Z$ and the contents of its $n \times d$ codebook matrix $B$, where each of the $n$ rows $B_k$ is a \enquote{memory block} vector. Hyperparameters $n$ and $d$ are determined empirically (see sections \ref{sec:num_memory_blocks} and \ref{sec:memory_block_size}). $Z$ is first partitioned evenly along its feature dimension into $s/d$ tensors, with each tensor $Z_f$ of size $d \times w \times h$. Each tensor $Z_f$ is further partitioned by spatial location into $w \cdot h$ vectors, denoted $Z_{f,x,y} \in \rm I\!R^{d}$, where $d$ is also the dimension of each row $B_k$ in the matrix $B$. For each vector $Z_{f,x,y}$ in $Z$, a similarity score $\gamma_k$ is calculated between it and each memory block $B_k$ as follows: 

\begin{equation}\label{equ:similarity}
\gamma_{f,x,y,k} = \langle Z_{f, x, y}, \frac{B_k}{{\| B_k \|}_2} \rangle
\end{equation}

Where $\langle \cdot, \cdot \rangle$ is the dot product, and ${\| \cdot \|}_2$ is the L2-norm. Because $B_k$ is normalized, $\gamma_{f,x,y,k}$ is highest for the memory block most similar in vector direction to the given $Z_{f,x,y}$. The memory block $B_k$ with the highest $\gamma$ similarity value replaces $Z_{f,x,y}$ at its corresponding location in $\widetilde{Z}$ as follows:

\begin{equation}\label{equ:select_feature_vector}
\widetilde{Z}_{f, x, y} \gets B_{k_{f,x,y}} \text{where } k_{f,x,y} = \underset{k}{\mathrm{argmax}}(\gamma_{f,x,y,k}) 
\end{equation}

Because $\widetilde{Z}$ is reconstructed entirely from memory blocks $B_k$, we can save all information needed to reconstruct $\widetilde{Z}$ again later by storing both $B$ and the values of $k$ at each $f,x,y$ location in $\widetilde{Z}$. Thus, the feature map for the $i^{th}$ training image can be stored as $m_i = (k_{1,1,1}, ..., k_{f, x, y}, ..., k_{s/d, w, h})$.

For example, in our main implementation, $Z$ is a $512 \times 13 \times 13$ tensor. $d = 16$ so that $Z$ is split into $32 \cdot 13 \cdot 13 = 5408$ vectors of length 16, which are each replaced in $\widetilde{Z}$ by a 16-dimensional memory block from a $256 \times 16$ matrix $B$. The memory blocks themselves occupy a near-negligible amount of memory: $256 \times 16 = 4096$ floating point values, compared to the $5408$ integers required to store a single compressed training example in this implementation.

\subsubsection{Training}

During training, both $Z$ and $\widetilde{Z}$ are passed separately through the classifier $P(\cdot)$ to obtain two classification probability vectors $p = P(Z)$ and $\widetilde{p} = P(\widetilde{Z})$, where the dimension of $p_t$ and $\widetilde{p}_t$ is equal to the total number of classes $C_t$ that have been seen in or before the current task $t$. The loss function $L$ used for training is a weighted sum of the cross-entropy losses $L_{CE}$ derived from $p$ and $\widetilde{p}$. With $y_c$ defined as the ground truth class label of a given image:

\begin{equation}\label{equ:loss_funtion}
L(p, \widetilde{p}, y_c) = \alpha L_{CE}(p,y_c) + \beta L_{CE}(\widetilde{p},y_c)
\end{equation}

Larger values of $\alpha$ penalize \enquote{direct} prediction errors from $P(Z)$, while larger values of $\beta$ penalize \enquote{codebook-out} prediction errors from $P(\widetilde{Z})$. Although our model generates class predictions based on both $Z$ and $\widetilde{Z}$, we use the empirically more accurate predictions from $Z$ during inference on the test set. Empirically, the best performance was achieved by including both direct and codebook-out predictions in the loss function for pretraining ($\alpha = 1, \beta = 1$), and then removing the direct loss for stream learning ($\alpha=0, \beta=1$) (see section~\ref{sec:analysis_loss}) for analysis). Setting $\alpha$ to 0 makes the loss function for new batches of images more similar to that used for replay, where only $\widetilde{Z}$ is available. Replacement of $Z$ by the reconstructed version $\widetilde{Z}$ can be viewed as both a means to efficiently mitigate catastrophic forgetting and a regularization technique to prevent overfitting and stabilize the CNN.

Although values in CRUMB's memory blocks play the role of activation values in their reconstruction of $\widetilde{Z}$, they are trainable parameters of the model. Backpropagation from $\widetilde{Z}$-based predictions generates gradients for the values in each memory block used for reconstruction, and stochastic gradient descent modifies the memory blocks towards minimizing the same training objective used for the network weights (cross-entropy loss). 

\begin{algorithm}[t]
\caption{CRUMB at task $t$}
   \label{algo:algorithm1}
\begin{algorithmic}
\footnotesize
   \STATE {\bfseries Input:} training images $I_t$ from new classes, stored codebook matrix $B$, replay buffer $X$ of stored memory block indices and their class labels (maximum number $n_X$ of total stored examples in $X$ varies by dataset). 
   \vspace{1mm}
   \STATE {\bfseries Training:}
   \FOR{batch {\bfseries in} $I_t$}
   \STATE Reconstruct feature map $Z$ as $\widetilde{Z}$ for each image in batch by concatenating memory blocks (rows of $B$)
   \STATE Train $P(\cdot)$ using loss $L(P(Z), P(\widetilde{Z}), y_c)$, with $\alpha=0$ in streaming
   \STATE Train memory blocks in $B$ that form part of $\widetilde{Z}$, using $L_{CE}(P(\widetilde{Z}),y_c)$
   \vspace{-3mm}
   \IF{$t > 1$}
   \STATE Randomly sample images $x$ out of $X$ to form a replay batch 
   \STATE Reconstruct $\widetilde{Z}$ for each $x$ by concatenating memory blocks
   \STATE Train $P(\cdot)$ using loss $L_{CE}(P(\widetilde{Z}),y_c)$
   \STATE Train memory blocks in $B$ that are part of $\widetilde{Z}$ via backpropagation
   \ENDIF
   \ENDFOR
   \STATE Store in X: memory block indices for reconstruction of every $j^{\text{th}}$ image
   \vspace{1mm}
  \STATE {\bfseries Testing:}
  \FOR{batch {\bfseries in} testing images}
  \STATE Compute predictions $p = P(F(\cdot))$ on test images using $Z$ only
  \ENDFOR
\end{algorithmic}
\end{algorithm}

\subsubsection{Initializing the codebook matrix and CNN}\label{subsection:matinitialize}
CRUMB's performance benefits from targeted initialization and pretraining of its CNN and the memory blocks in its codebook matrix, especially in the class-instance setting. The values in the codebook matrix $B$ directly replace those in \enquote{natural} feature maps derived from images during training - accordingly, $B$ is initialized using a simple univariate distribution designed to match that of natural feature maps from a pretrained network. In early experiments, we tried initializing the codebook using k-means cluster centers from feature vectors of CIFAR100 images \cite{krizhevsky2009learning}, but this did not improve performance. 

Stream learning performance was substantially improved by pretraining CRUMB on ImageNet \cite{deng2009imagenet} classification with 1000 classes, as compared to applying CRUMB to stream learning with a CNN pretrained by standard methods. Pretraining tunes the values in the memory blocks, and also regularizes the CNN in preparation for stream learning by training it to make predictions using lossy reconstructions of feature maps. 

\subsubsection{Replay to mitigate catastrophic forgetting}\label{subsection:mitigatef}

In online stream learning (see section~\ref{sec:protocols}), the model is presented with images $I_t$ from new classes $c^{new}$ in task $t$, where $c^{new}$ are drawn from the subset of classes in the dataset that the model has not seen in previous tasks. 

Replay of examples from previous tasks is a proven strategy to mitigate catastrophic forgetting in class-incremental settings \cite{rebuffi2017icarl, wu2019large, chaudhry2018efficient, aljundi2019gradient}, and feature-level replay can be considerably more memory-efficient than storing raw images \cite{hayes2020remind}. We store compressed representations of feature maps from images in each task, and then replay a batch of stored feature maps after each batch of new images during later tasks to mitigate forgetting. 

CRUMB stores up to $n_X$ pairs of labels and tensors $(y_i, m_i)$, corresponding to images from old classes $c^{old}$ of previous tasks. Depending on the number of seen classes $C_{t-1}$, the storage for each old class contains $n_X/C_{t-1}$ pairs. $n_X$ is chosen for each dataset depending roughly on the total number of classes. 
Some algorithms select representative image examples to store and replay based on different scoring functions \cite{chen2012super,koh2017understanding,Brahma_2018_CVPR_Workshops,ho2023prototype}. However, random sampling uniformly across classes yields outstanding performance in continual learning tasks \cite{wen2018few}, and we adopt this strategy to select examples from the buffer for replay. To choose training examples for storage in the replay buffer, CRUMB keeps every $j^{\text{th}}$ image in each batch, where $j$ is calculated by dividing the number of training images in each task by the replay buffer capacity $n_x$, such that the buffer is filled near the end of the current task's training epoch. In the class-instance setting, this maximizes sample diversity by minimizing the number of frames sampled from within the same video clip, and further avoiding sampling frames from the same clip that are in close temporal proximity. 

For replay-based baseline methods (iCARL, REMIND, etc), we limit the number of examples that can be stored in the buffer to fit within a memory budget that is fixed across all methods (see details in supplementary Section S5). Aside from $n_x$ and the training batch size (which is smaller for video datasets), CRUMB uses the same hyperparameters for all datasets, including $n$, $d$, $\alpha$, $\beta$, learning rate, and initialization and pretraining protocols. 

\setlength{\textfloatsep}{0pt}

\section{Results}\label{result}

\begin{table*}
\begin{center}
\caption{\textbf{CRUMB outperforms state-of-the-art algorithms on most benchmarks.} Each number is the mean top-1 accuracy on all tasks/classes after the completion of stream learning. Values are averages from 10 (CORe50, Toybox, iLab, iCub, iLab+CORe50, Online-CIFAR100) or 5 (Online-ImageNet) independent runs. The highest accuracy in each column (excluding the offline upper bound) is in bold, while the second-highest is italicized. 
Algorithm name abbreviations can be found in section~\ref{sec:baselines}. Class-instance, in which video frames are presented in temporal order, is only applicable to video datasets CORe50, Toybox, iLab, iCub, and iLab+CORe50. Due to resource constraints, for Online-ImageNet, iCub, and CORe50+iLab, we tested a subset of baseline algorithms. Class-instance and class-i.i.d. settings are abbreviated as ``inst.'' and ``i.i.d.'' respectively. Results are grouped vertically with memory distillation and replay methods at the top (Ours - CoPe), weight regularization methods in the middle (EWC - LwF), and lower/upper bounds at the bottom. Data preparation methods are detailed in supplementary Section S4.A.
}
\begin{tabular}{l|cc|cc|cc|cc|cc|c|c}
\textbf{} & \multicolumn{2}{c|}{CORe50 \cite{lomonaco2017core50}} & \multicolumn{2}{c|}{Toybox \cite{wang2018toybox}} & \multicolumn{2}{c|}{iLab \cite{borji2016ilab}} & \multicolumn{2}{c|} {iCub \cite{pasquale2016object}} & \multicolumn{2}{c|}{iLab+CORe50 } & CIFAR100 \cite{krizhevsky2009learning} & ImageNet \cite{deng2009imagenet} \\
\textbf{} & inst. & i.i.d. & inst. & i.i.d. & inst. & i.i.d. & inst. & i.i.d. & inst. & i.i.d. & i.i.d. & i.i.d. \\ \hline
Ours & \textbf{78.5} & \textbf{81.2} & \textbf{74.9} & \textit{75.7} & \textbf{77.9} & \textit{79.5} & \textbf{60.0} & \textbf{65.8} & \textbf{66.0} & \textbf{74.4} & \textbf{49.9} & \textbf{48.9} \\
\rowcolor[HTML]{DDDDDD} 
REMIND \cite{hayes2020remind} & \textit{77.0} & \textit{76.0} & \textit{66.2} & \textbf{84.1} & \textit{48.1} & \textbf{81.0} & \textit{33.2} & \textit{58.5} & \textit{22.9} & \textit{67.2} & \textit{38.2} & \textit{46.2} \\ 
iCARL \cite{rebuffi2017icarl} & 27.0 & 28.5 & 27.3 & 26.5 & 15.6 & 23.6 & 23.4 & 20.9 & 17.8 & 23.2 & 15.9 & 18.5 \\
\rowcolor[HTML]{DDDDDD} 
GEM \cite{lopez2017gradient} & 11.9 & 13.5 & 14.3 & 15.7 & 13.0 & 12.8 & 5.2 & 6.1 & 4.6 & 4.5 & 3.5 & 2.9 \\
RM \cite{bang2021rainbow} & 12.0 & 12.4 & 9.8 & 20.8 & 18.2 & 9.3 & - & - & - & - & 4.2 & - \\
\rowcolor[HTML]{DDDDDD} 
AAN \cite{liu2021adaptive} & 14.0 & 15.6 & 13.2 & 17.6 & 10.6 & 15.0 & - & - & - & - & 6.6 & - \\
GSS \cite{aljundi2019gradient} & 15.0 & 15.6 & 14.7 & 15.0 & 13.0 & 12.8 & - & - & - & - & 3.2 & - \\
\rowcolor[HTML]{DDDDDD} 
BiC \cite{wu2019large} & 10.2 & 11.8 & 11.0 & 10.2 & 11.2  & 10.9 & - & - & - & - & 4.0 & - \\
CoPE \cite{de2020continual} & 16.6 & 16.3 & 21.7 & 22.4 & 17.6 & 18.6 & - & - & - & - & 8.8 & - \\
\hline

\rowcolor[HTML]{DDDDDD} 
EWC \cite{kirkpatrick2017overcoming} & 12.2 & 12.4 & 14.3 & 15.7 & 13.5 & 13.0 & 7.3 & 6.4 & 11.7 & 12.0 & 3.9 & 0.1 \\
MAS \cite{aljundi2018memory} & 14.4 & 17.4 & 18.9 & 19.2 & 20.5 & 22.1 & 4.8 & 4.8 & 4.4 & 4.3 & 5.5 & 0.1 \\
\rowcolor[HTML]{DDDDDD} 
SI \cite{zenke2017continual} & 12.0 & 12.9 & 14.3 & 15.5 & 12.8 & 13.0 & 5.3 & 5.7 & 4.4 & 5.0 & 3.6 & 8.8 \\
Stable SGD \cite{mirzadeh2020understanding} & 13.7 & 13.2 & 13.5 & 13.8 & 9.8 &  6.9 & - & - & - & - & 7.3 & - \\

\rowcolor[HTML]{DDDDDD} 
LwF \cite{li2017learning} & 12.5 & 12.4 & 21.9 & 20.9 & 10.5 & 11.9 & - & - & - & - & 4.2 & - \\
\hline 

Lower bound & 12.1 & 12.8 & 15.5 & 16.9 & 12.8 & 16.4 & 5.8 & 6.0 & 4.5 & 4.6 & 3.5 & 3.0 \\
\rowcolor[HTML]{DDDDDD} 
Upper bound & 85.3 & 84.6 & 91.0 & 92.0 & 91.3 & 91.4  & 76.9 & 78.0 & 83.1 & 81.7 & 69.0 & 56.1
\end{tabular} \label{tab:all_results}
\end{center}
\end{table*}

\begin{figure*}[ht]
    \centering
    \begin{minipage}{\textwidth}
        \includegraphics[width=5.1cm, height = 4.0cm]{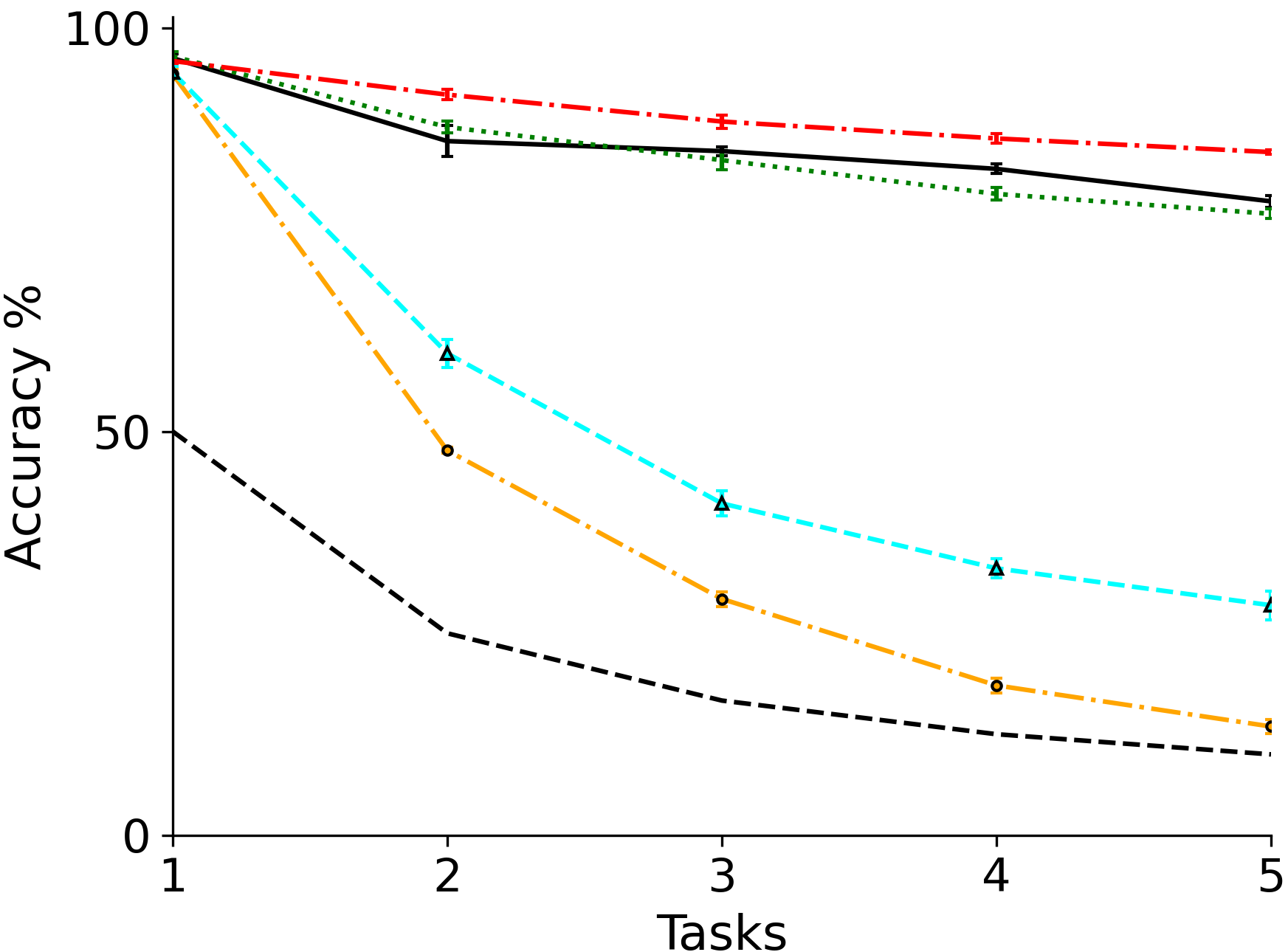}\hfill
        \includegraphics[width=5.3cm, height = 4.0cm]{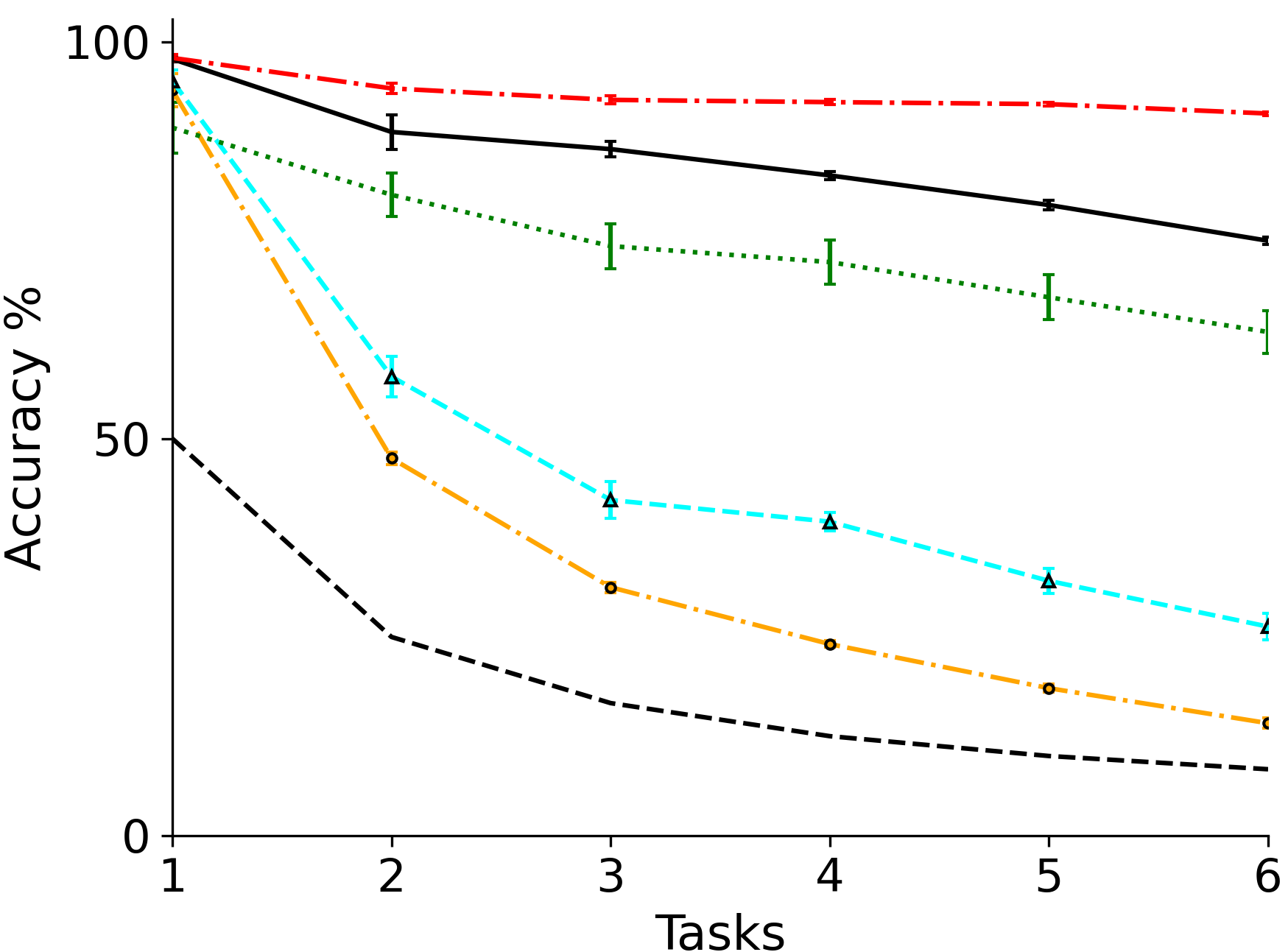} \hfill
        \includegraphics[width=5.6cm, height = 4.0cm, trim = 10mm 1.5mm 0 0]{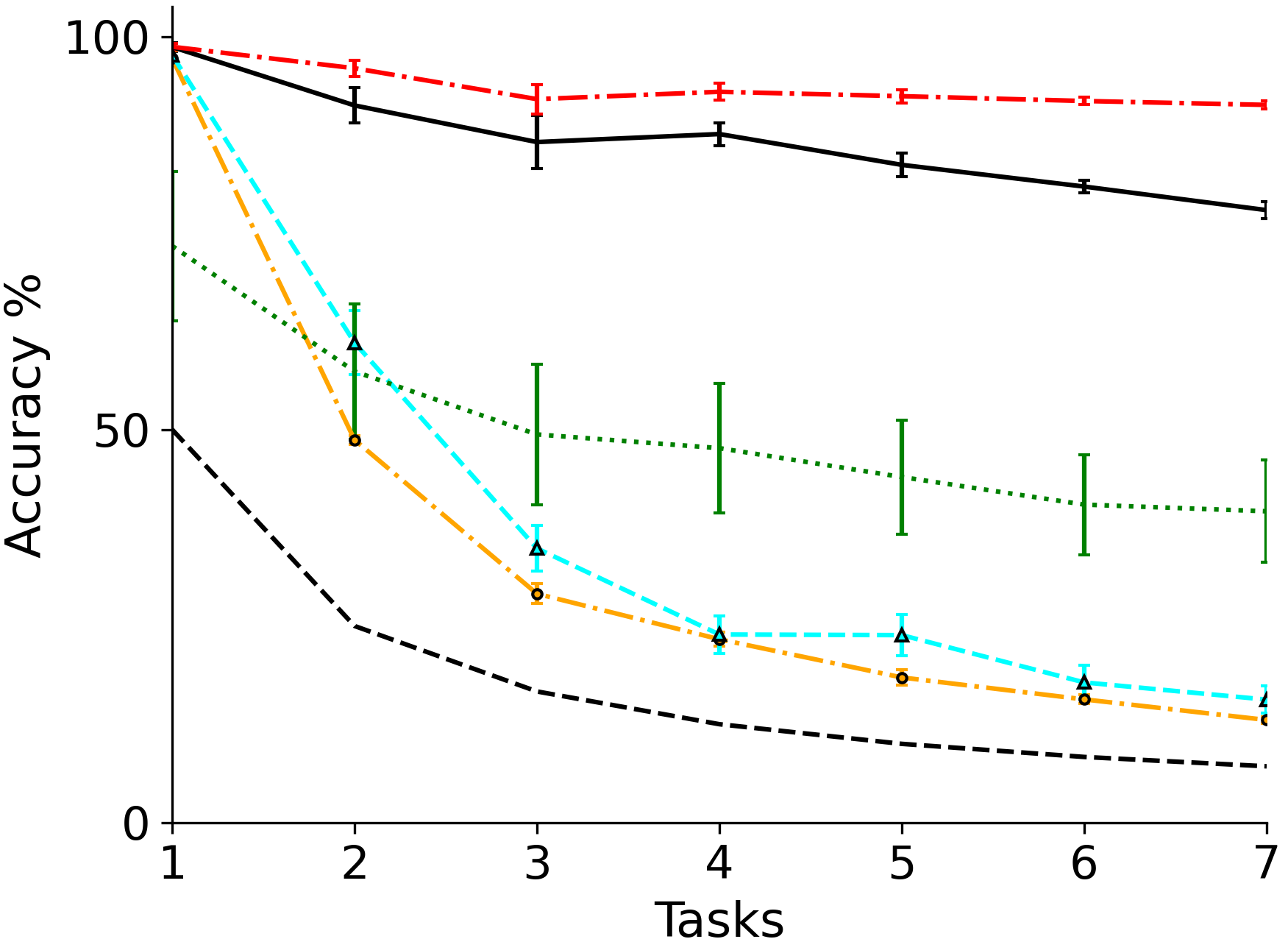}  \hfill
        
    \hspace{1.0cm} a. CORe50 (class-instance) \hspace{2.0cm} b. Toybox (class-instance) \hspace{2.2cm} c. iLab  (class-instance)
    \\
    \end{minipage}\hfill

    \vspace{-2mm}

    \begin{minipage}{\textwidth}
        \includegraphics[width=6.4cm, height = 4.3cm]{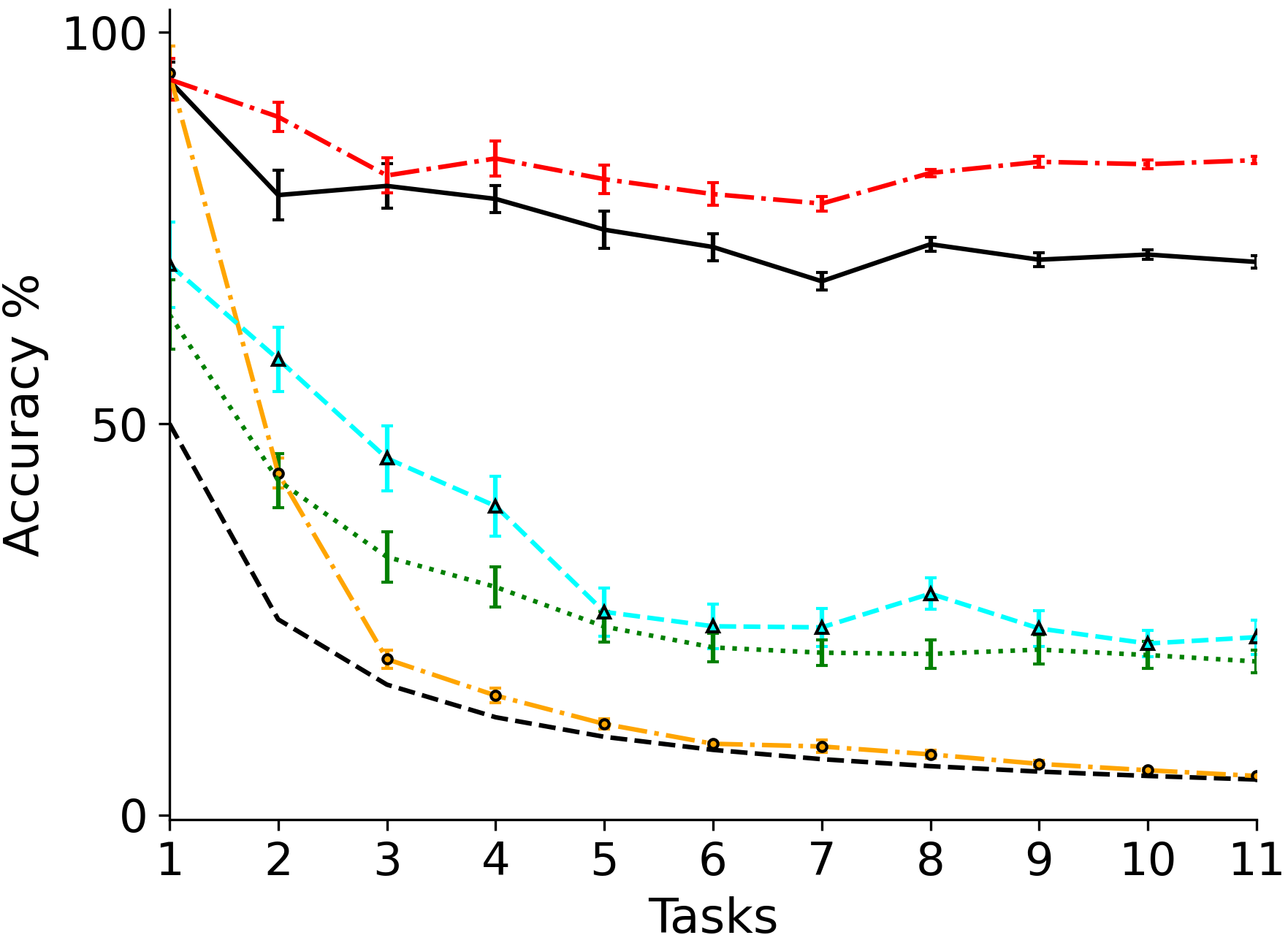}\hfill
        \includegraphics[width=6.4cm, height = 4.3cm]{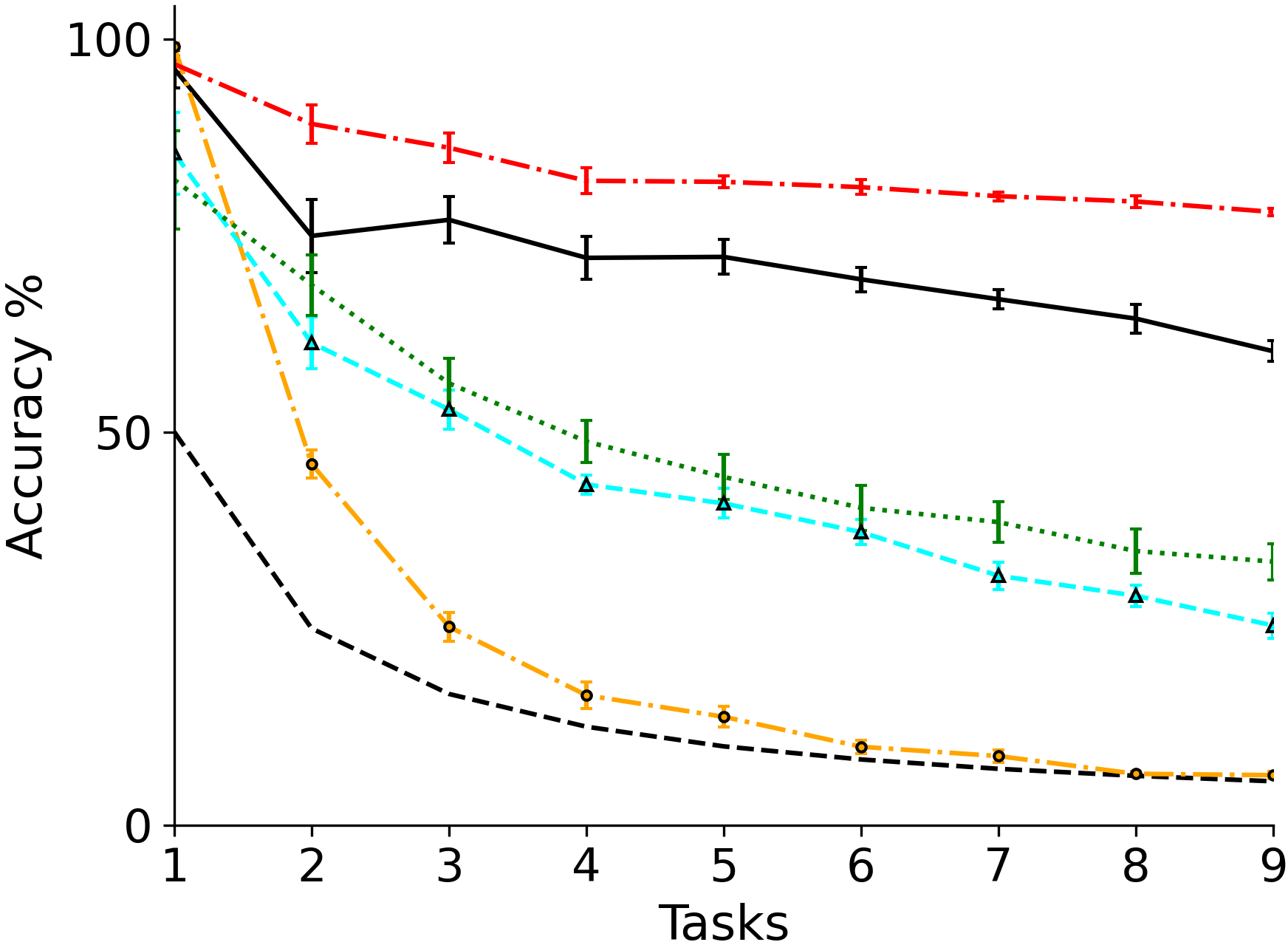} \hfill
        \includegraphics[height = 3.0cm]{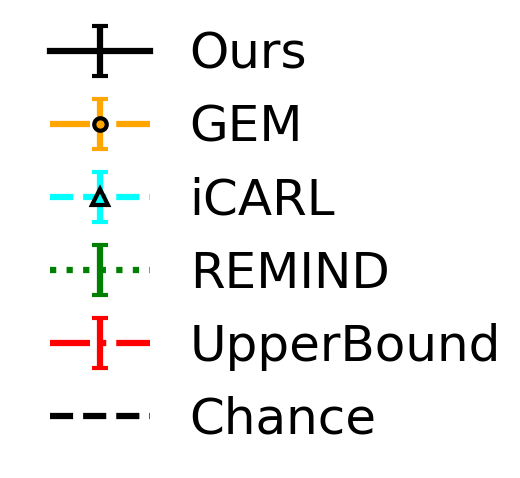}\hfill
        
    \hspace{1.2cm} d. iLab + CORe50 (class-instance)  \hspace{3.0cm} e. iCub  (class-instance) \hspace{2.3cm} f. Legend  
    \\
    \end{minipage}\hfill

    \vspace{-3mm}
    
    \begin{minipage}{\textwidth}
       
       \includegraphics[width=7.5cm, height = 5.3cm]{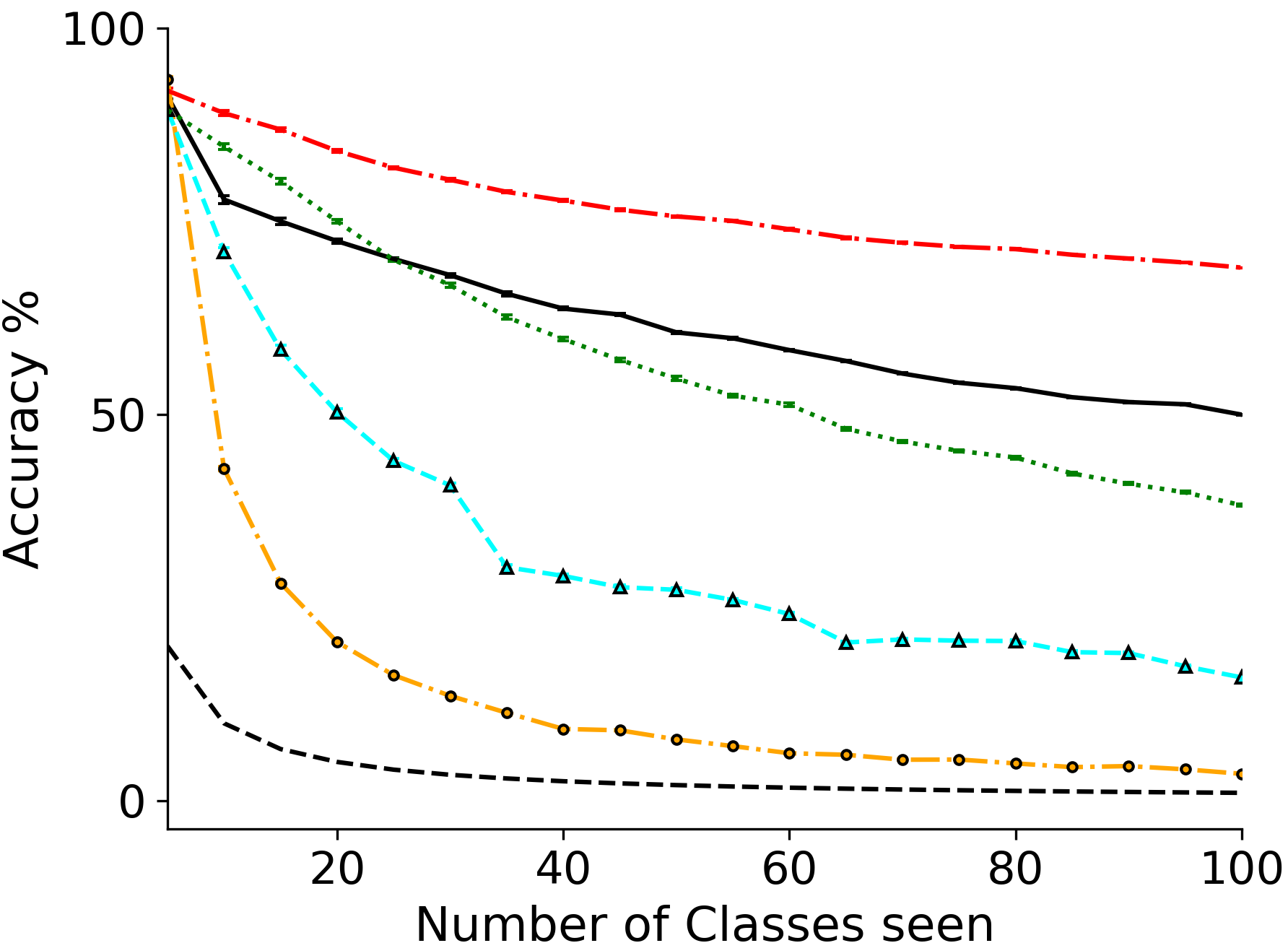}\hfill
        \includegraphics[width=7.5cm, height = 5.3cm]{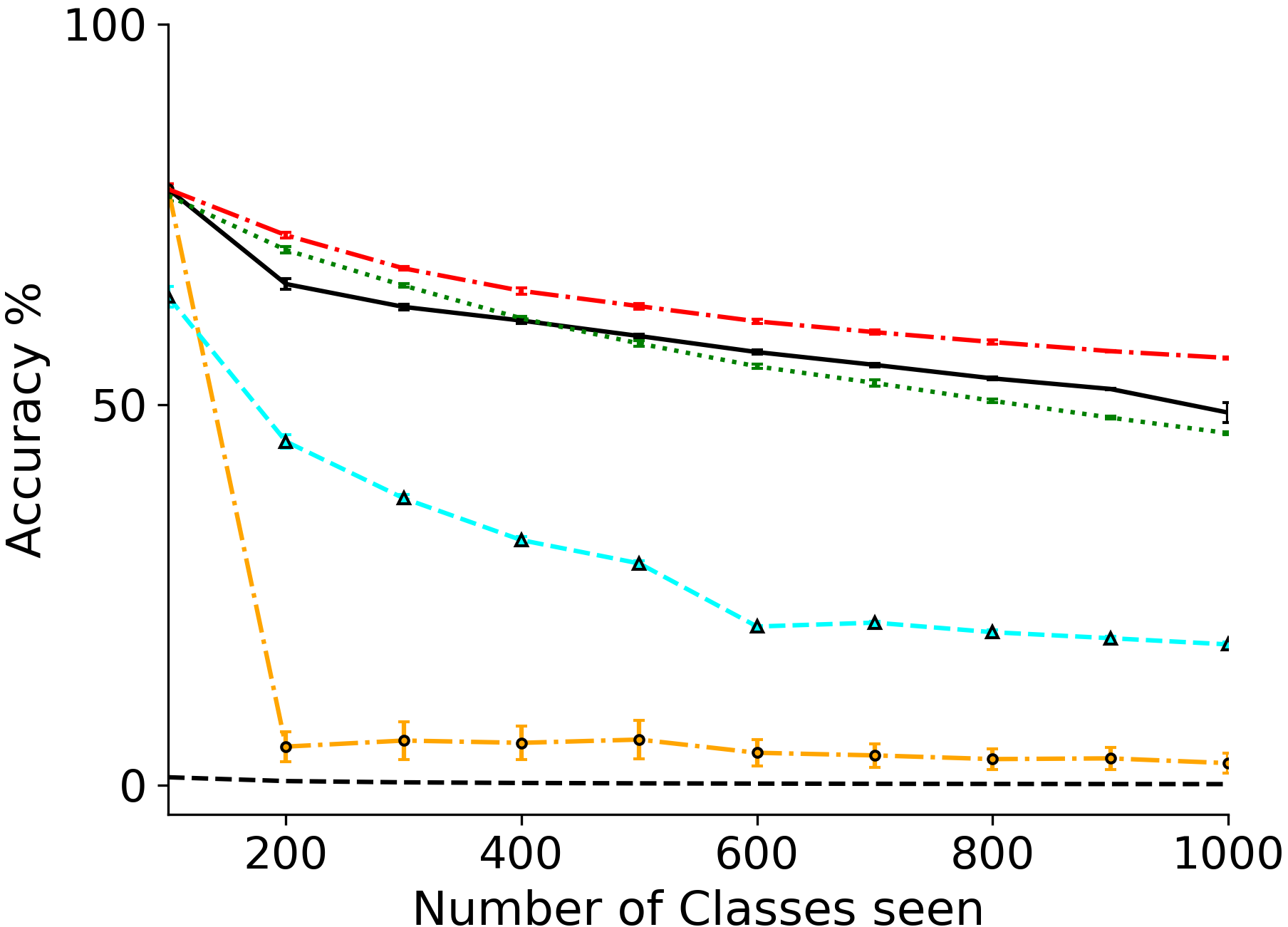}  \hfill

    \vspace{0.1cm} \hspace{2cm} g. Online-CIFAR100 (class-i.i.d.) \hspace{5.5cm} h. Online-ImageNet (class-i.i.d.)
    \end{minipage}\hfill

    \caption{\textbf{CRUMB outperforms most baseline algorithms and approaches the upper bound on some datasets}. Line plots show top-1 accuracy in online stream learning on video datasets (a) CORe50 (b) Toybox (c) iLab (d) iLab + CORe50 (e) iCub in the class-instance training protocol (class-i.i.d. plots are in supplementary Fig. S1), as well as image datasets (g) Online-CIFAR100 and (h) Online-ImageNet (class-i.i.d.). All models train on the first task for many epochs, but \textbf{view each image only once} on all subsequent tasks. Accuracy estimates are the mean from 10 runs (5 runs for ImageNet), where each run has different class and image/video clip orderings.
    Error bars show the root-mean-square error (RMSE) among runs. Results for all baselines are in Table~\ref{tab:all_results}.}
\label{fig:video_streaming_main}
\end{figure*}

\subsection{Stream learning on video datasets}
A naive CNN trained on stream learning benchmarks learns each task effectively, but rapidly and catastrophically forgets all prior tasks in doing so. In contrast, a brute-force approach to overcoming catastrophic forgetting that achieves excellent performance in a stream learning setting is to store all encountered images and corresponding class labels, shuffle them, and exhaustively retrain on the resulting dataset in an offline, i.i.d. fashion. This renders the benchmark equivalent to offline class-incremental learning (\enquote{upper bound} in Fig.~\ref{fig:video_streaming_main}) \cite{prabhu2020gdumb}. By storing a subset of old examples and using a compositional strategy to compress these examples, CRUMB allows CNNs to approach the performance of a brute-force approach with roughly an order-of-magnitude reduction in training time and a tiny 0.013\% fraction (on CORe50) of the memory footprint. Accordingly, given a fixed memory budget, CRUMB outperforms all competing models in all five tested video stream learning datasets in the class-instance setting, often by large margins. For example, as shown in Fig.~\ref{fig:video_streaming_main}a-e, CRUMB's top-1 accuracy on all tasks after class-instance stream learning exceeds that of REMIND by 1.5\%, 8.7\%, 29.8\%, 26.8\%, and 43.1\%, iCARL by 51.5\%, 47.6\%, 62.3\%, 36.6\%, and 48.2\%, and GEM by 66.6\%, 60.6\%, 64.9\%, 54.8\%, and 61.4\% on CORe50, Toybox, iLab, iCub and iLab+CORe50 respectively. The class-instance performance of all models is shown in Table~\ref{tab:all_results}. CRUMB approaches the offline upper bound to within 6.8\%, 16.1\%, 13.4\%, 16.9\% and 17.1\% on the same datasets, demonstrating strong mitigation of catastrophic forgetting.  

The less challenging class-i.i.d. setting is similar to class-instance in that tasks are learned sequentially without revisiting previous tasks, and that each image is seen by the model only once. However, all images within each class-i.i.d. task are shuffled in an i.i.d. manner, removing the local temporal correlations introduced by sequential frames in video clips. As with class-instance, CRUMB achieves excellent class-i.i.d. performance: as shown in supplementary Fig.~S1a-e, CRUMB's top-1 accuracy on all tasks after class-i.i.d. learning exceeds that of iCARL by 52.7\%, 49.2\%, 55.9\%, 44.9\%, and 51.2\% and of GEM by 67.7\%, 60\%, 66.7\%, 59.7\%, and 69.9\% on CORe50, Toybox, iLab, iCub and iLab+CORe50 respectively. The class-i.i.d. performance of all models is shown in Table~\ref{tab:all_results}. CRUMB approaches the offline upper bound to within 3.4\%, 16.3\%, 11.9\%, 12.2\% and 7.3\% on the same datasets. The performance of REMIND and CRUMB was comparable on class-i.i.d., with CRUMB's accuracy higher than REMIND's by 5.2\%, 7.3\% and 7.2\% on CORe50, iCub and iLab+CORe50 respectively. However, REMIND's accuracy was higher than CRUMB's by 8.4\% and 1.5\% on Toybox and iLab respectively. 

On all benchmarks, CRUMB's closest competitor by far was REMIND, with all other methods exhibiting much lower accuracy. In general, the regularization baselines greatly underperformed the replay-based methods. This is perhaps due to limited exposure to each task given that each image may be visited only once, and/or because of overfitting to temporally correlated data, especially in the class-instance setting: replay addresses both of these issues while regularization methods generally do not. Because we used a fixed memory budget for replay methods, CRUMB is able to store many more examples than replay methods based on raw images, such as iCARL and GEM. This increases the diversity of the replayed stimuli.  

\subsection{Stream learning on natural image datasets}
Although stream learning of CORe50, Toybox, iLab, iCub, and iLab+CORe50 is highly challenging, these datasets have only 10-24 classes each. To demonstrate CRUMB's capacity for long-range stream learning of many classes, we employed standard image datasets Online-CIFAR100 and Online-ImageNet. CRUMB outperformed all baselines on both of these datasets (see Table~\ref{tab:all_results}). On Online-CIFAR100, CRUMB's mean top-1 accuracy after class-i.i.d. stream learning exceeds that of REMIND by 11.7\%, iCARL by 34\%, and GEM by 46.4\%, performing within 19.1\% of the offline upper bound. On Online-Imagenet, CRUMB outperforms REMIND by 2.7\%, iCARL by 30.4\%, and GEM by 46\%, performing within 7.2\% of the offline upper bound (see also Fig.~\ref{fig:video_streaming_main}g-h).  

\subsection{Memory and runtime efficiency}

\begin{table}[]
\begin{center}
\caption{CRUMB uses only 3.7-4.1\% of REMIND's peak RAM usage, and its runtime is approximately 15-43\% of REMIND's. 
}
\begin{tabular}{|l|ll|ll|}
\hline
\multirow{2}{*}{\textbf{Dataset}} & \multicolumn{2}{l|}{\textbf{Peak RAM (GB)}}            & \multicolumn{2}{l|}{\textbf{Runtime (hours)}}        \\ \cline{2-5} 
                                  & \multicolumn{1}{l|}{\textbf{Ours}}  & \textbf{REMIND} & \multicolumn{1}{l|}{\textbf{Ours}} & \textbf{REMIND} \\ \hline
\textbf{CIFAR100}                 & \multicolumn{1}{l|}{\textbf{0.036}} & 0.87           & \multicolumn{1}{l|}{\textbf{0.29}} & 1.91            \\ \hline
\textbf{Imagenet}                 & \multicolumn{1}{l|}{\textbf{1.66}}  & 44.34           & \multicolumn{1}{l|}{\textbf{15.50}} & 35.64           \\ \hline
\end{tabular} \label{tab:memory_runtime_compare}
\end{center}
\end{table}

CRUMB's closest competitor in top-1 accuracy is REMIND \cite{hayes2020remind}. Both models require specific pretraining procedures. REMIND trains a product quantizer using k-means clustering of feature vectors, which requires a large portion of training data to be held in memory simultaneously. In contrast, CRUMB's codebook matrix is trained by backpropagation in tandem with CNN parameter updates. This approach requires only 3.7-4.1\% of the peak RAM usage of REMIND for large datasets such as Online-CIFAR100 and Online-Imagenet. CRUMB also has a runtime only about 15-43\% as long as REMIND's (Table~\ref{tab:memory_runtime_compare}). 

\subsection{Model analysis} \label{sec:model_analysis}

To elucidate the importance of CRUMB's various components, we performed a series of ablation studies and experiments with altered training procedures. Accuracy results on CORe50 in both class-instance and class-i.i.d. settings are included for each experiment in Tables~\ref{tab:crumb_beats_images} and \ref{tab:network_analysis}, but throughout the text in this section we discuss class-instance results except where otherwise stated. Experiment names are in \textbf{bold} throughout this section. We conducted statistical significance tests for each experiment (see supplementary Section S4.B). 

\subsubsection{Replay: \textit{n} CRUMB feature maps beat \textit{n} images} \label{sec:crumb_beats_images}

\begin{table}
\begin{center}
\caption{\textbf{CRUMB performs better on video stream learning with \textit{n} feature map representations in its replay buffer (\textmd{Ours}) than with \textit{n} raw images (\textmd{Im. replay}), even though the former uses only 3.6\% as much memory}. This finding is demonstrated for both class-i.i.d. and class-instance, across a range of buffer sizes on CORe50, and across 5 video datasets, although it does not persist for image datasets Online-CIFAR100 and Online-ImageNet. For CORe50 in the \enquote{Dataset} column, the buffer size is indicated as $n_x$. The table shows mean final top-1 accuracy on all tasks, averaged across 5 independent runs that each begin with an independent pretraining run. Significant differences from \textbf{Ours} are marked with *}
\begin{tabular}{|l|l|c|c|}
\hline \textbf{Dataset} & \textbf{Experiment} & \textbf{Class-inst.} & \textbf{Class-i.i.d.} \\ \hline
\multirow{2}{*}{CORe50 ($n_x=200$)}  
& Ours & 78.22 & 79.93 \\
& Im. replay & 67.80* & 75.88* \\ \hline

\multirow{2}{*}{CORe50 ($n_x=400$)}  
& Ours & 79.08 & 81.37 \\
& Im. replay & 71.55* & 78.36* \\ \hline

\multirow{2}{*}{CORe50 ($n_x=800$)}  
& Ours & 79.46 & 82.21 \\
& Im. replay & 68.73* & 81.67 \\ \hline

\multirow{2}{*}{CORe50 ($n_x=1600$)}  
& Ours & 79.30 & 82.60 \\
& Im. replay & 69.64* & 79.62* \\ \hline

\multirow{2}{*}{CORe50 ($n_x=3200$)}  
& Ours & 81.39 & 80.83 \\
& Im. replay & 70.50* & 79.72 \\ \hline

\multirow{2}{*}{CORe50 ($n_x=6400$)}  
& Ours & 79.39 & 83.76 \\
& Im. replay & 69.60* & 80.55* 
\\ \hline
\multirow{2}{*}{Toybox}  
& Ours & 68.19 & 68.91 \\
& Im. replay & 64.03* & 62.68* \\ \hline
\multirow{2}{*}{iLab}  
& Ours & 67.80 & 71.96 \\
& Im. replay & 52.36* & 63.38* \\ \hline
\multirow{2}{*}{iCub}  
& Ours & 58.33 & 63.02 \\
& Im. replay & 50.50* & 47.88* \\ \hline
\multirow{2}{*}{iLab+CORe50}  
& Ours & 61.89 & 72.55 \\
& Im. replay & 59.05* & 67.58* \\ \hline
\multirow{2}{*}{Online-CIFAR100} 
& Ours & - & 47.97 \\
& Im. replay & - & 47.81 \\ \hline
\multirow{2}{*}{Online-ImageNet} 
& Ours & - & 23.99 \\
& Im. replay & - & 42.27* \\ \hline
\end{tabular} \label{tab:crumb_beats_images}
\end{center}
\end{table}

\hspace{12pt}Feature-level replay is the main mechanism by which CRUMB prevents catastrophic forgetting. Removing replay dramatically reduces accuracy by 64.9\%. However, CRUMB does not require storing a large number of feature maps to mitigate forgetting: reducing buffer size $n_X$ from 200 (\textbf{Ours}) to 100 (\textbf{Half capacity}) and to 50 (\textbf{Quarter capacity}) had a relatively small impact, with 4.7\% and 13.3\%  accuracy drops respectively.

The quality of stored replay examples is also important. \textbf{Ours}, which stores memory block indices to compositionally reconstruct up to $n_X$ feature maps, had dramatically higher accuracy than storing \textit{the same number $n_X$ of entire raw images} and training the network without any feature map reconstruction (\textbf{Image replay}), even though CRUMB's reconstruction of feature maps inevitably discards information and uses only 3.6\% as much memory. As shown in Table~\ref{tab:crumb_beats_images}, \textbf{Ours} attains 10.4\% higher accuracy than \textbf{Image replay} on CORe50 (buffer size $n_x=200$), 4.2\% higher on Toybox, 15.4\% higher on iLab, 7.8\% higher on iCub, and 2.8\% higher on iLab+CORe50. This result appears to hold only for the five video streaming datasets, however: \textbf{Ours} attained accuracy 0.16\% higher than \textbf{Image replay} on Online-CIFAR100 (not statistically significant), and 18.3\% lower than \textbf{Image replay} on Online-ImageNet, in the class-i.i.d. setting. \textbf{Ours} uses only 3.6\% as much memory as \textbf{Image replay}: when the amount of memory usage is held constant (e.g., in gigabytes) instead of the maximum number of replay buffer items $n_x$, CRUMB outperforms image replay methods such as iCARL by very large margins on all datasets (see Table~\ref{tab:all_results}). 

Table~\ref{tab:crumb_beats_images} also shows that CRUMB continues to outperform image replay on CORe50 when memory resources for the replay buffer are not constrained. Even at $n_x=6400$, where algorithms can store the entire CORe50 training set in the replay buffer, \textbf{Ours} outperforms \textbf{Image replay} by 9.8\% on class-instance and 3.2\% on class-i.i.d. The reduced performance of \textbf{Image replay} relative to \textbf{Ours} is partly rescued by adding CRUMB pretraining (\textbf{Ours p.t. + im. rep.}, 3.7\% and 2.2\% below \textbf{Ours} in class-instance and class-i.i.d. respectively), even though the memory blocks play no role in either of the two image replay conditions during streaming (see Table \ref{tab:network_analysis}). 

Replaying high-level features also contributed to CRUMB's performance. Storing $n_X$ low-level feature maps from layer 3 instead of layer 12 (\textbf{Early feature replay} vs. \textbf{Ours}) reduced performance by 16.6\%. CRUMB effectively stores memories with a higher level of abstraction than both \textbf{Image replay} and \textbf{Early feature replay}, and comes with both accuracy and memory efficiency improvements. The Memory Recall method (MeRec, \cite{zhang2021memory}) uses a further level of abstraction by storing only the element-wise mean and standard deviation of feature activations for each learned class, and generating examples for replay by sampling from a Gaussian distribution parameterized by these values. In \textbf{MeRec replay}, we implemented MeRec's replay mechanism as a drop-in replacement for CRUMB's memory block reconstruction at the same network layer, and observed accuracy 56.9\% and 41.3\% below \textbf{Ours} for class-instance and class-i.i.d. respectively, but 8.1\% and 28.1\% above \textbf{No replay}. MeRec stores an amount of data equivalent to two complete feature maps (mean and standard deviation) per class. In the implementation of CRUMB used for \textbf{Ours}, this is equivalent to 16 compressed feature maps per class or $n_x=160$ in total, which is fewer than \textbf{Ours} at $n_x=200$ but more than \textbf{Half capacity} at $n_x=100$.  

\subsubsection{CRUMB pretraining primes CNN weights for streaming} \label{sec:pretraining_primes}

Our model's performance is maximized by using CRUMB to pretrain the CNN and memory blocks on ImageNet prior to stream learning, particularly in the class-instance condition. Using randomly initialized memory blocks and a CNN pretrained without CRUMB (\textbf{Vanilla pretrain}) instead of CRUMB pretraining (\textbf{Ours}) reduced performance by 25.5\% and 3.2\% on class-instance and class-i.i.d. respectively. Our results also indicate that the benefit of CRUMB pretraining is attributable primarily to changes in the CNN weights, rather than changes to the memory blocks. Starting stream learning with CRUMB-pretrained weights and randomly re-initialized memory blocks (\textbf{Pretrain weights}) performs only 1.1\% and 0.5\% worse than \textbf{Ours}, while starting with vanilla-pretrained weights and CRUMB-pretrained memory blocks (\textbf{Pretrain mem. blocks}) is 23.7\% and 2.8\% worse than \textbf{Ours}, a marginal improvement over vanilla pretraining. As explained in section \ref{sec:crumb_beats_images}, CRUMB pretraining also improves stream learning performance when raw images are used for replay. 

CRUMB pretraining using the smaller CIFAR100 dataset (100 classes) instead of ImageNet (1000 classes) (\textbf{CIFAR100 pretrain}) decreases accuracy by 11.6\%. 

In \textbf{Freeze memory}, no updates to memory blocks were allowed after pretraining. This had no statistically significant effect on accuracy, indicating that fine-tuning the memory blocks was unnecessary for stream learning on CORe50. 

\subsubsection{CRUMB pretraining induces a shape bias in the CNN} \label{sec:shape_bias}

\begin{table}
\begin{center}
\caption{\textbf{Ablation and other experiments demonstrate the importance of CRUMB's various components.} Top-1 accuracy on all tasks after stream learning is averaged over 5 runs for all experiments. * denotes significant difference from \textbf{Ours} ($p < 0.01$, paired-samples t-tests on batches of 100 images).}
\footnotesize
\begin{tabular}{|l|lll}
\hline
\textbf{Category} & \multicolumn{1}{l|}{\textbf{Experiment name}} & \multicolumn{1}{l|}{\textbf{\begin{tabular}[c]{@{}l@{}}Class-inst.\\ \% avg. \\ accuracy\end{tabular}}} & \multicolumn{1}{l|}{\textbf{\begin{tabular}[c]{@{}l@{}}Class-i.i.d. \\ \% avg. \\ accuracy\end{tabular}}} \\ \hline
Unablated                                                                               & \multicolumn{1}{l|}{\textbf{Ours}}              & \multicolumn{1}{l|}{\textbf{78.22}} & \multicolumn{1}{l|}{\textbf{79.93}}          \\ \hline
\multirow{4}{*}{Replay format}                                                          
& \multicolumn{1}{l|}{Image replay}               & \multicolumn{1}{l|}{67.80*} & \multicolumn{1}{l|}{75.88*}                          \\ \cline{2-4} 
& \multicolumn{1}{l|}{Ours p.t. + im. rep.}     & \multicolumn{1}{l|}{74.49*} & \multicolumn{1}{l|}{77.72*}                          \\ \cline{2-4} 
& \multicolumn{1}{l|}{Early feature replay}       & \multicolumn{1}{l|}{61.64*} & \multicolumn{1}{l|}{64.28*}                          \\ \cline{2-4} 
& \multicolumn{1}{l|}{MeRec replay \cite{zhang2021memory}}               & \multicolumn{1}{l|}{21.35*} & \multicolumn{1}{l|}{38.65*}                          \\ \hline 

\multirow{3}{*}{Replay ablation}                                                        
& \multicolumn{1}{l|}{Half capacity}              & \multicolumn{1}{l|}{73.55*} & \multicolumn{1}{l|}{75.14*}                          \\ \cline{2-4} 
& \multicolumn{1}{l|}{Quarter capacity}           & \multicolumn{1}{l|}{64.90*} & \multicolumn{1}{l|}{67.80*}                          \\ \cline{2-4} 
& \multicolumn{1}{l|}{No replay}                  & \multicolumn{1}{l|}{13.28*} & \multicolumn{1}{l|}{10.58*}                          \\ \hline

\multirow{4}{*}{Pretraining ablation}  
& \multicolumn{1}{l|}{Vanilla pretrain}               & \multicolumn{1}{l|}{52.70*} & \multicolumn{1}{l|}{76.69*}                          \\ \cline{2-4}
& \multicolumn{1}{l|}{Pretrain weights}          & \multicolumn{1}{l|}{77.08*} & \multicolumn{1}{l|}{79.40}                          \\ \cline{2-4} 
& \multicolumn{1}{l|}{Pretrain mem. blocks}     & \multicolumn{1}{l|}{54.54*} & \multicolumn{1}{l|}{77.18*}                          \\ \cline{2-4} 
& \multicolumn{1}{l|}{CIFAR100 pretrain}          & \multicolumn{1}{l|}{66.64*} & \multicolumn{1}{l|}{74.96*}                          \\ \hline

\multirow{1}{*}{Freeze memory}
& \multicolumn{1}{l|}{Freeze memory}              & \multicolumn{1}{l|}{78.06}  & \multicolumn{1}{l|}{80.44}                           \\ \hline

\multirow{3}{*}{\begin{tabular}[c]{@{}l@{}}Memory block\\ init.\end{tabular}}           
& \multicolumn{1}{l|}{Normal init.}              & \multicolumn{1}{l|}{47.70*} & \multicolumn{1}{l|}{74.28*}                          \\ \cline{2-4} 
& \multicolumn{1}{l|}{Uniform init.}              & \multicolumn{1}{l|}{39.84*} & \multicolumn{1}{l|}{64.82*}                          \\ \cline{2-4} 
& \multicolumn{1}{l|}{Dense matched init.}        & \multicolumn{1}{l|}{77.24} & \multicolumn{1}{l|}{78.84*}                          \\ \hline

\multirow{8}{*}{\begin{tabular}[c]{@{}l@{}}Number of\\ memory blocks\end{tabular}}      
& \multicolumn{1}{l|}{1 block}                    & \multicolumn{1}{l|}{9.60*}  & \multicolumn{1}{l|}{9.47*}                           \\ \cline{2-4} 
& \multicolumn{1}{l|}{2 blocks}                   & \multicolumn{1}{l|}{64.28*} & \multicolumn{1}{l|}{71.23*}                          \\ \cline{2-4} 
& \multicolumn{1}{l|}{4 blocks}                   & \multicolumn{1}{l|}{70.10*} & \multicolumn{1}{l|}{75.77*}                          \\ \cline{2-4} 
& \multicolumn{1}{l|}{8 blocks}                   & \multicolumn{1}{l|}{74.60*} & \multicolumn{1}{l|}{79.96}                           \\ \cline{2-4} 
& \multicolumn{1}{l|}{16 blocks}                  & \multicolumn{1}{l|}{77.70} & \multicolumn{1}{l|}{80.30}                           \\ \cline{2-4} 
& ...                                             & ...                        & ...                                                   \\ \cline{2-4} 
& \multicolumn{1}{l|}{256 blocks \textbf{(Ours})} & \multicolumn{1}{l|}{\textbf{78.22}} & \multicolumn{1}{l|}{\textbf{79.93}}         \\ \cline{2-4} 
& \multicolumn{1}{l|}{512 blocks}                 & \multicolumn{1}{l|}{78.71}          & \multicolumn{1}{l|}{79.74}                  \\ \hline
\
\multirow{6}{*}{\begin{tabular}[c]{@{}l@{}}Memory block\\ size\end{tabular}}           
& \multicolumn{1}{l|}{4-dim. blocks}              & \multicolumn{1}{l|}{74.21*}         & \multicolumn{1}{l|}{77.55*}                 \\ \cline{2-4}
& \multicolumn{1}{l|}{8-dim. blocks (\textbf{Ours})} & \multicolumn{1}{l|}{\textbf{78.22}} & \multicolumn{1}{l|}{\textbf{79.93}}      \\ \cline{2-4} 
& \multicolumn{1}{l|}{16-dim. blocks}             & \multicolumn{1}{l|}{79.05*}         & \multicolumn{1}{l|}{81.02*}                 \\ \cline{2-4}
& \multicolumn{1}{l|}{32-dim. blocks}             & \multicolumn{1}{l|}{78.17}         & \multicolumn{1}{l|}{79.93}                 \\ \cline{2-4}
& \multicolumn{1}{l|}{16-dim. blocks adj.}        & \multicolumn{1}{l|}{79.69*}         & \multicolumn{1}{l|}{82.36*}                 \\ \cline{2-4} 
& \multicolumn{1}{l|}{32-dim. blocks adj.}        & \multicolumn{1}{l|}{80.31*}         & \multicolumn{1}{l|}{81.64*}                  \\ \hline      

\multirow{3}{*}{Loss functions}                                                         
& \multicolumn{1}{l|}{Ours - direct loss}         & \multicolumn{1}{l|}{73.82*} & \multicolumn{1}{l|}{78.11*}                          \\ \cline{2-4} 
& \multicolumn{1}{l|}{Ours + direct loss}         & \multicolumn{1}{l|}{65.40*} & \multicolumn{1}{l|}{69.20*}                          \\ \cline{2-4} 
& \multicolumn{1}{l|}{Direct loss}                & \multicolumn{1}{l|}{48.12*} & \multicolumn{1}{l|}{50.07*}                          \\ \hline

\multirow{1}{*}{\begin{tabular}[c]{@{}l@{}}CNN Architecture\end{tabular}} 
& \multicolumn{1}{l|}{MobileNetV2 CNN}        & \multicolumn{1}{l|}{76.13*} & \multicolumn{1}{l|}{79.27}                          \\ \hline

\end{tabular} \label{tab:network_analysis}
\end{center}
\end{table}

\begin{figure}[t]
    \centering
    \includegraphics[width=\linewidth]{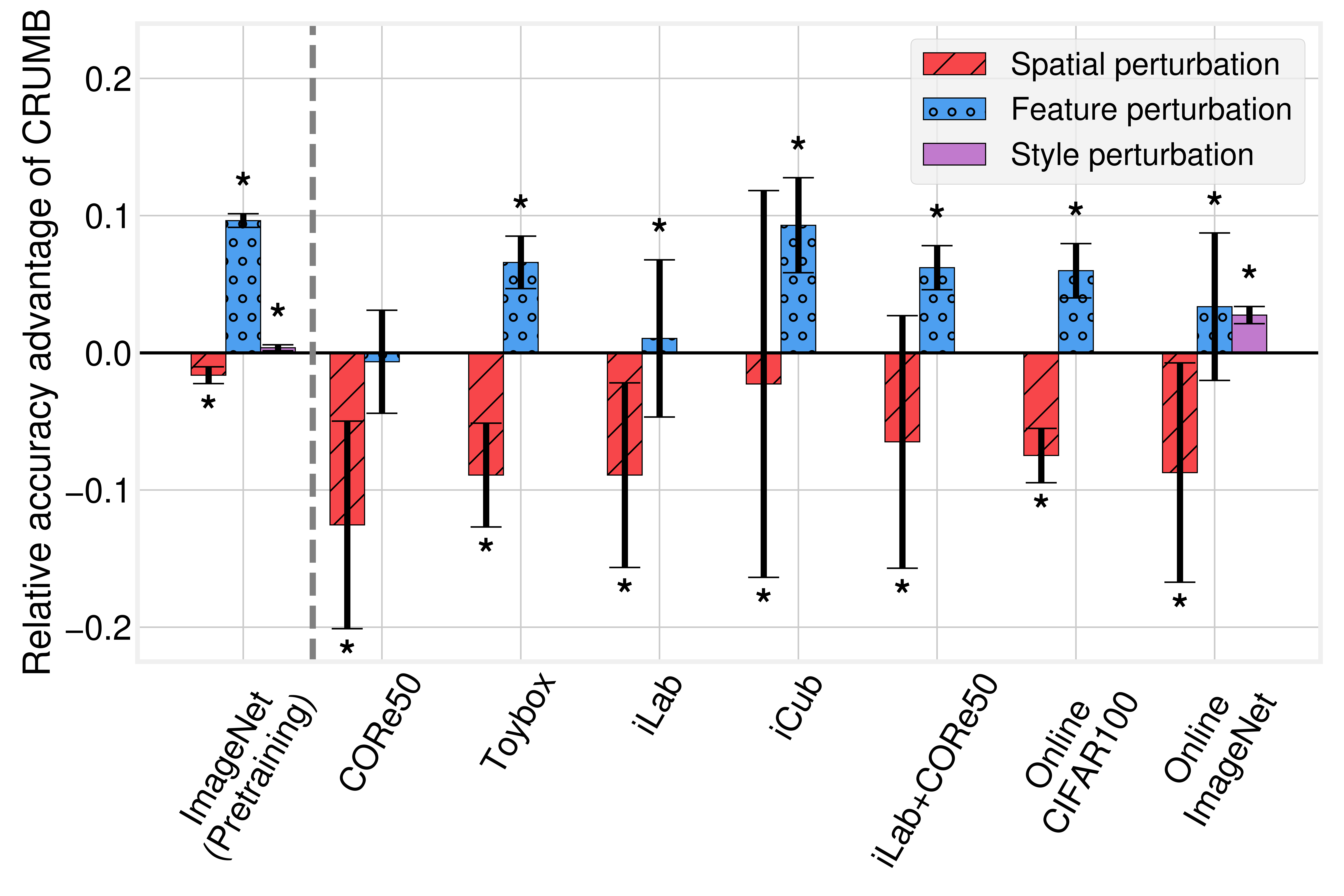}
    \vspace{-8mm}
    \caption{\textbf{CRUMB pretraining induces a bias towards shape information that often persists through stream learning}. The height of each bar shows how much smaller (or larger, if negative) CRUMB's drop in normalized test set accuracy under a perturbation is, in comparison to a control network (see section \ref{sec:pretraining_primes}). ``Spatial perturbation'' shuffles the spatial positions of all feature vectors in an intermediate feature map (at the same layer where it is reconstructed by CRUMB), ``feature perturbation'' randomly sets half of the feature map's features to zero, and \enquote{style perturbation} uses images from Stylized-ImageNet \cite{geirhos2018}. Streaming results (to the right of grey dotted line) are in the class-instance setting for the video datasets and class-i.i.d. for CIFAR100 and ImageNet. Error bars are standard errors of the mean of relative accuracy advantage among 5 (CIFAR100 and ImageNet) or 10 (other datasets) independent runs. * denotes a statistically significant difference from 0, as determined by a Wilcoxon signed-rank test (see supplementary Section S4.B). 
    }
	\label{fig:shape_bias_class_instance}
\end{figure}

We hypothesized that CRUMB's pretraining procedure induces a bias towards shape information over texture information by training the CNN to make class predictions using lossy feature representations with unperturbed spatial distributions. A bias towards shape information has been shown to help mitigate forgetting by flattening the local minima for each task in the loss landscape \cite{shi2022robustness}. To gauge CRUMB's degree of reliance on shape and texture information, we evaluated its performance on test set examples with three different perturbations. ``Spatial perturbation'' and ``feature perturbation'' modify the $13 \times 13 \times 512$ feature map at the same level where it is reconstructed using memory blocks. In ``spatial perturbation,'' the positions of all of the 512-dimensional feature vectors are randomly shuffled in the $13 \times 13$ spatial grid, destroying global shape information but leaving feature information intact. In ``feature perturbation,'' for each image independently we set a random selection of 50\% of the features (i.e., 256 out of 512 total features) to zero, perturbing feature information but leaving coarse shape information intact. Spatial and feature perturbations do not directly interact with CRUMB's reconstruction mechanism, because CRUMB uses the original, non-reconstructed feature map to make class predictions. The ``feature perturbation'' strategy assumes that feature information is more related to texture information than shape information. Therefore, we also include ``style perturbation'' for ImageNet by testing on the the Stylized-ImageNet dataset, which consists of images with heavily distorted local textures but largely intact global object shapes \cite{geirhos2018}. We would expect the performance of a relatively shape-biased network to be more severely reduced by spatial perturbation and less severely affected by feature and style perturbations than a control network. Fig.~\ref{fig:shape_bias_class_instance} shows CRUMB's ``relative accuracy advantage'' for each perturbation across several datasets. The relative accuracy advantage is calculated by dividing the accuracy drop (unperturbed accuracy minus perturbed accuracy) for each perturbation by the network's unperturbed accuracy, and subtracting this value for CRUMB from the corresponding value for a control network. The ``ImageNet (pretraining)'' condition in Fig.~\ref{fig:shape_bias_class_instance} shows CRUMB's shape bias on ImageNet following pretraining on ImageNet, with lower resilience against spatial perturbations (red bars with diagonal lines) and higher resilience against feature (blue with circles) and style (plain purple) perturbations. The control network is pretrained on ImageNet using the same procedure but without the CRUMB feature reconstruction step, thereby using only the ``direct'' loss for parameter updates. 
The other conditions in Fig.~\ref{fig:shape_bias_class_instance} correspond to CRUMB models evaluated on the test sets of these datasets after stream learning in class-instance (CORe50, Toybox, iLab, iCub, and iLab+CORe50) or class-i.i.d. (Online-CIFAR100 and Online-ImageNet) settings. Class-i.i.d. results for the video datasets are shown in supplementary Fig. S3. Here, the control network performs stream learning with raw-image replay and ``direct'' loss instead of CRUMB's feature-level replay and ``codebook-out'' loss. Although shape bias testing results are noisy on some of the datasets, CRUMB most often retains a degree of bias towards shape information over texture/feature information after the completion of stream learning. 

\subsubsection{CRUMB can learn with very few memory blocks} \label{sec:num_memory_blocks}

\hspace{12pt}CRUMB's performance did not change dramatically with changes to the number of memory blocks. Reducing the memory block count from \textbf{256 blocks} to as few as \textbf{16 blocks}, which effectively shrinks the library of feature combinations available to reconstruct feature maps, did not significantly decrease accuracy. Reducing the count further to \textbf{8 blocks} decreased accuracy by 3.6\%, and reducing to \textbf{4} or \textbf{2 blocks} decreased accuracy by 8.1\% and 13.9\% respectively. Increasing to \textbf{512 blocks} did not significantly increase accuracy. This suggests a saturation effect, where a relatively small number of memory blocks is sufficient to reconstruct a wide variety of feature maps. 

\subsubsection{CRUMB is robust to different memory block sizes, and memory block size affects memory efficiency} \label{sec:memory_block_size}

CRUMB performs well with a range of memory block sizes. Decreasing the number of elements in each memory block from 8 to 4 (\textbf{4-dim.~blocks}) results in a modest decrease in performance, 4.0\% and 2.4\% on class-instance and class-i.i.d. respectively. Increasing the number of elements from 8 to 16 or 32 (\textbf{16-dim.~blocks}, \textbf{32-dim.~blocks}), which arguably makes accurate reconstruction of feature maps more challenging because higher-dimensional vectors must be replaced by discrete choices of memory blocks, had negligible impact on performance (see Table~\ref{tab:network_analysis}). 

The maximum number of examples stored in CRUMB's replay buffer ($n_x$) was held constant for the memory block size perturbations above. However, increasing the dimension of the memory blocks from 8 to 16 or 32 means that only half or one-quarter as many blocks respectively are needed to reconstruct each feature map, so only half/one-quarter as many indices need to be stored in the replay buffer per image. This allows double/quadruple the number of examples to be stored in the replay buffer within the same memory budget. When we allowed the maximum number of examples stored in the buffer to change accordingly ($2n_x$ for \textbf{16-dim.~blocks~adj.}, $4n_x$ for \textbf{32-dim.~blocks~adj.}), we observed accuracy improvements: \textbf{16-dim.~blocks~adj.} achieves 1.5\% and 2.4\% higher accuracy than \textbf{Ours} ($n_x$ with 8-dimensional blocks) on class-instance and class-i.i.d. respectively, and \textbf{32-dim.~blocks~adj.} achieves 2.1\% and 1.7\% higher accuracy. During hyperparameter tuning for our main results, we observed that 16-dimensional memory blocks maximized testing accuracy. 

\subsubsection{Loss from reconstructed features is sufficient} \label{sec:analysis_loss}

\hspace{12pt}CRUMB's performance is affected by the choice of components in its loss function. The loss function (equation~\ref{equ:loss_funtion}) is the weighted sum of two terms, \enquote{direct loss} and \enquote{codebook-out loss.} 
Our experiments show that the best performance is achieved when both direct loss and codebook-out loss are included in pretraining, but only codebook-out loss is included during stream learning. Removing direct loss from pretraining (\enquote{\textbf{Ours - direct loss}}) results in a 4.4\% drop in accuracy in the later stream learning tasks - learning from only reconstructed feature maps from start to finish, including during pretraining, is sufficient for decent performance. Including only codebook-out loss (\enquote{\textbf{Ours}}) in stream learning yields a dramatic 30.1\% gain in accuracy compared to using only direct loss (\enquote{\textbf{Direct loss}}), and a gain of 12.8\% compared to using a weighted sum of direct loss and codebook-out loss (\enquote{\textbf{Ours + direct loss}}), despite the fact that only the direct, non-reconstructed feature map is used for inference on the test set. 

\subsubsection{Initialization of the memory blocks matters}

CRUMB's performance is somewhat sensitive to the initialization of the values in the memory blocks. CRUMB trains its memory blocks in tandem with network weights after initialization, and concatenates them in different combinations to reconstruct feature maps produced by an intermediate network layer. We compared stream learning performance of four memory block initialization strategies, including initializing with values drawn from (1) a standard normal distribution (\textbf{Normal init.}), (2) a uniform distribution on the interval $[0, 1]$ (\textbf{Uniform init.}), (3) a distribution designed to match that of the non-zero values in the feature maps to be reconstructed, with 64\% of all values reset to zero to approximately match the sparsity of typical feature maps (\textbf{Ours}), and (4) the same as (3), but with no values set to zero (\textbf{Dense matched init.}). Accuracy for \textbf{Normal init.} was 30.5\% and 5.7\% lower than \textbf{Ours} for class-instance and class-i.i.d. protocols respectively, accuracy for \textbf{Uniform init.} was 38.4\% and 15.1\% lower, and accuracy for \textbf{Dense matched init.} was 1.0\% ($p = 0.045 > 0.01$) and 1.1\% lower (see Table~\ref{tab:network_analysis}). It appears that drawing initial values for the memory blocks from a similar distribution to that of natural feature maps improves performance. 
When applying CRUMB to new network architectures, a simple alternative procedure to initialize the memory blocks would be to obtain feature maps from a batch of images, pool all values from all feature maps into one long vector, and initialize each memory block value by randomly sampling a value from this vector. 

\subsubsection{CRUMB is applicable across CNN architectures}
In \textbf{MobileNetV2 CNN}, we implement CRUMB using the MobileNetV2 \cite{sandler2018mobilenetv2} CNN backbone instead of SqueezeNet \cite{iandola2016squeezenet}. Here, CRUMB reconstructs the $14 \times 14 \times 64$ input to MobileNetV2's sixth layer, using 256 8-dimensional memory blocks as in \textbf{Ours}. With the total memory usage of the replay buffer held constant, performance of \textbf{MobileNetV2 CNN} is comparable to \textbf{Ours} with only a 2.1\% accuracy drop in class-instance and a 0.7\% (not significant) drop in class-i.i.d. 

\subsubsection{Some memory blocks are coarsely interpretable} \label{sec:visualizations}

\begin{figure}[t]
    \centering
    \includegraphics[width=\linewidth]{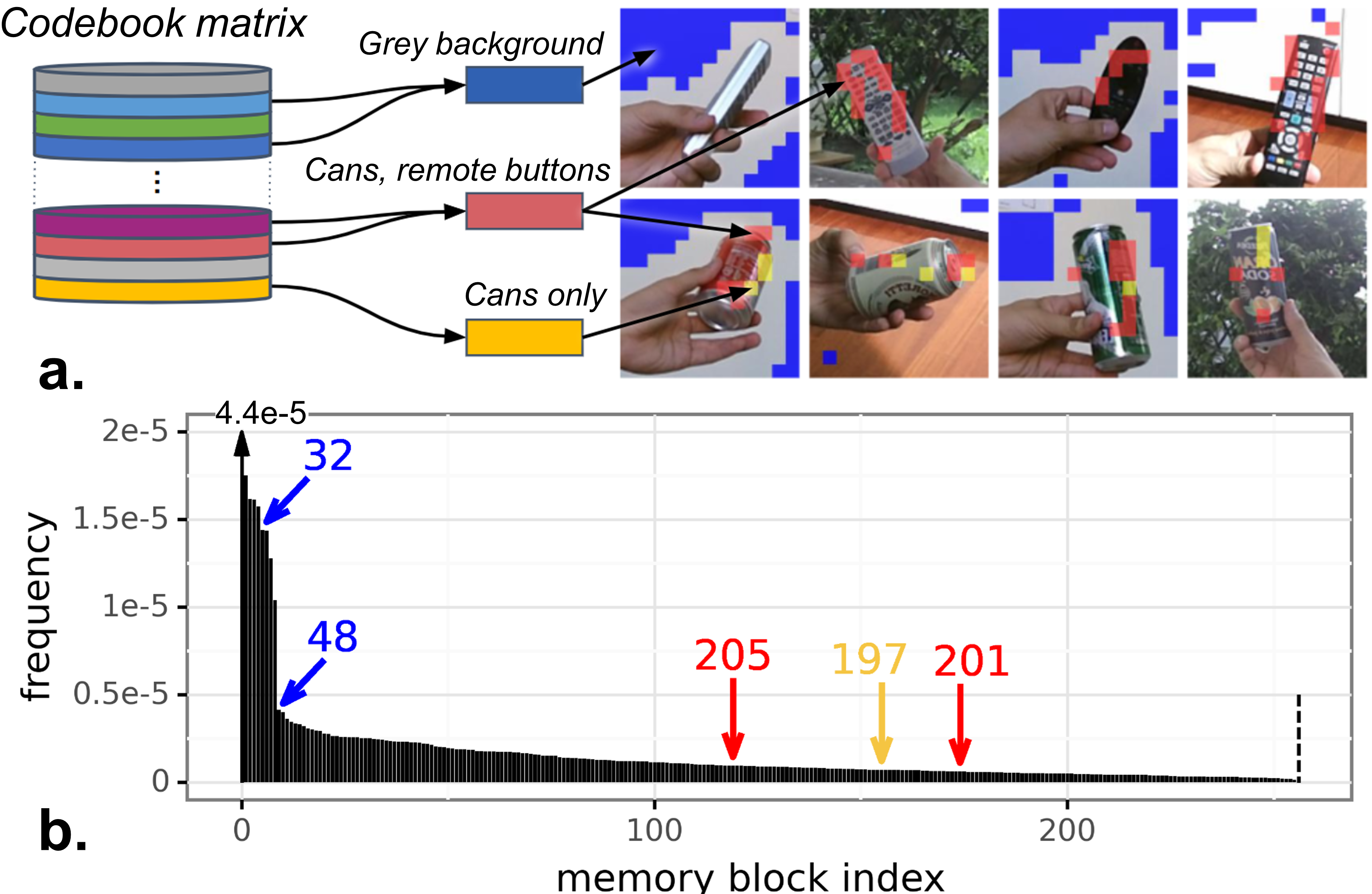}
    \vspace{-8mm}
    \caption{\textbf{Some memory blocks appear to have semantic interpretations}. Panel \textbf{a} shows images of \enquote{remote controls} and \enquote{cans} in the CORe50 test set, showing all-or-none activation of specific memory blocks at corresponding image locations. Of the 256 memory blocks in the codebook, blocks with indices 32 and 48 (blue squares) both similarly respond to greyish background regions, but not bright white or other backgrounds. Blocks 201 and 205 (red) both respond to buttons on remote controls and features of drink cans, while block 197 (yellow) responds only to can features. Similar blocks are aggregated by color (for blue and red) to produce a clearer visualization. Panel \textbf{b} shows the sorted usage frequencies in the CORe50 test set of each of the 256 memory blocks. Colored arrows show the blocks visualized in panel \textbf{a}. The upward black arrow shows the most-used block with frequency 4.4e-5.}
	\label{fig:can_remote_visualizations}
\end{figure}

Visualizations of image locations where specific memory blocks are activated (Fig.~\ref{fig:can_remote_visualizations}) show that some memory blocks appear to be human-interpretable. Some blocks responded to features seen in images of one specific class or of a subset of classes, and others responded to features that are likely irrelevant to classification. In addition to the blocks visualized in Fig.~\ref{fig:can_remote_visualizations}, we found blocks that tend to respond to vertical lines, crosshatch patterns on balls and cups, pure white backgrounds, vegetation backgrounds, and wooden floor backgrounds, each of which can be interpreted as a semantic, compositional part of various test set images. Given the observations earlier in this section that either randomly re-initializing memory blocks prior to stream learning (\textbf{Pretrain weights}) or freezing memory blocks during stream learning (\textbf{Freeze memory}) has minimal effects on performance, it is not necessarily the case that these interpretable associations indicate learned representations within the memory blocks themselves. Another possibility is that a sufficient diversity of memory blocks allows useful associations between memory blocks and features or classes to be learned by the CNN through changes to the network weights. 

The procedure for generating the visualizations in panel \textbf{a} of Fig.~\ref{fig:can_remote_visualizations} can be understood as follows. For this analysis, we used a CRUMB model trained on CORe50 in the class-instance setting. CORe50 test set images are first passed through the early layers of the CNN to produce a feature map, which CRUMB then reconstructs by concatenating memory block vectors to produce an approximated version of the original feature map (see section~\ref{subsection:reconstructmem}). In this study, each feature map is of size $13 \times 13 \times 512$, meaning spatial dimensions of $13 \times 13$ with 512 features at each spatial location. Each memory block is one of 256 row vectors in the $256 \times 8$ codebook matrix used for this analysis. The memory blocks are 8-dimensional vectors, so each spatial location in the feature map's $13 \times 13$ grid is represented by a 512-dimensional vector formed by concatenating $512/8 = 64$ memory blocks end-to-end. Color coding in Fig.~\ref{fig:can_remote_visualizations}a shows at most one block per spatial location, the one activated by the first 8 features in the 512-dimensional feature vector, even though 64 memory blocks are activated at each location in total. We focus on the first 8 features for visualization purposes, because it is not necessarily the case that blocks activated by the first set of 8 features encode the same image features as they might when activated by the $k^{th}$ set of 8 features (where $2 \leq k \leq 64$). To produce the images in Fig.~\ref{fig:can_remote_visualizations}a, each test set image is divided into a square $13 \times 13$ grid. Image grid locations are overlaid with colored squares, such that the color of each square depends on the memory block activated by the first 8 features at the corresponding spatial location in the feature map reconstructed by CRUMB. We only assigned colors to a handful of memory blocks with interesting properties, and we assigned the same color to sets of memory blocks that seemed to respond to very similar features. Fig.~\ref{fig:can_remote_visualizations}b shows the sorted distribution of the frequencies with which each of the 256 memory blocks were used to reconstruct feature maps from the CORe50 test set, with color and memory block index-coded arrows indicating the memory blocks visualized in Fig.~\ref{fig:can_remote_visualizations}a. 

\section{Discussion and conclusion}\label{sec:discussion}

We developed a novel compositional replay strategy to tackle the problem of online stream learning, in which algorithms must learn tasks incrementally from non-repeating, temporally correlated inputs. Our algorithm, CRUMB, learns a set of \enquote{memory blocks} that are selected via cosine similarity and concatenated to reconstruct feature maps from an intermediate CNN layer. The indices of selected memory blocks are stored for a subset of training images, enabling memory-efficient replay of feature maps to mitigate catastrophic forgetting. CRUMB achieves state-of-the-art online stream learning accuracy across 7 datasets. Furthermore, CRUMB outperforms replay of an equal number of raw images by large accuracy margins across 5 video datasets, despite using only 3.6\% as much memory as image replay.

Several factors seem to make important contributions to CRUMB's high performance. As shown in Fig.~\ref{fig:shape_bias_class_instance}, pretraining with memory block reconstruction biases the CNN towards attending to object shapes rather than textures, an effect that typically endures throughout stream learning and which has been demonstrated to reduce catastrophic forgetting by flattening the loss minima for each task \cite{shi2022robustness}. This could explain why CRUMB pretraining improves performance even if raw image replay is used during stream learning (``Ours p.t. + im. rep.'' in Table~\ref{tab:network_analysis}), in which case feature map reconstruction plays no role whatsoever during stream learning. 

Backpropagation updates to memory blocks during pretraining and stream learning appear to be less important for performance than updates to the CNN weights. Randomly re-initializing the memory blocks after pretraining has a small negative effect on performance only in the class-instance setting, while keeping only the pretrained memory blocks but resetting to the original ``vanilla'' pretrained CNN weights before stream learning has a much larger negative impact (``pretrain weights'' vs. ``pretrain mem. blocks'' in table~\ref{tab:network_analysis}). Furthermore, freezing the memory blocks during streaming has no discernible effect (``freeze memory'' in table ~\ref{tab:network_analysis}). However, ``normal init.'' and ``uniform init.'' in Table~\ref{tab:network_analysis} show that the choice of probability distribution used to randomly initialize the memory blocks can have a dramatic impact on performance, with best performance attained by matching the univariate distribution of the memory blocks to that of natural feature maps. It seems to be important for stream learning that the later layers of the network receive feature maps with consistent univariate statistics, whether they are natural feature maps from the feature extractor layers or reconstructed feature maps from memory block concatenation. 

Experiments in table~\ref{tab:network_analysis} also show that the design of CRUMB's loss function is important. When training on new images, using only \enquote{codebook-out loss} from classification on reconstructed feature maps leads to much less forgetting than using \enquote{direct loss} from natural feature maps, either together with codebook-out loss (``ours + direct loss'') or in isolation (``direct loss''). Only codebook-out loss is available when replaying feature maps reconstructed from memory blocks: using only codebook-out loss for new images means that only codebook-out loss is used throughout stream learning, rather than switching between direct loss for new examples and codebook-out loss for replayed examples. It appears that CRUMB's memory blocks form a shared, discretized basis for encoding training examples in feature space, which has a stabilizing effect on the CNN during stream learning. It is also notable in this context that, unlike during streaming, the inclusion of direct loss is important for gradient updates during pretraining (``Ours - direct loss'' has lower accuracy than ``Ours'' in table~\ref{tab:network_analysis}). Combined with the observation that memory blocks should ideally be initialized from a univariate distribution approximating that of natural feature maps, this suggests that the pretraining process helps the CNN align its processing of natural and reconstructed feature maps, enabling the network to learn only from reconstructed feature maps during streaming even while continuing to make its most accurate predictions using natural feature maps instead.  

The hypothesis that CRUMB's memory blocks provide a shared feature-level basis that stabilizes the CNN is consistent with the observation that, although CRUMB pretraining improves performance of raw image replay (``Ours p.t. + im. rep.'' in table ~\ref{tab:network_analysis}), it still does not match CRUMB's performance with the replay buffer size $n_x$ held constant: we speculate that redundant pixel-level information in raw images introduces additional noisy variation into the distribution of feature maps, which affects network stability. Another observation consistent with this hypothesis is that CRUMB only outperforms raw image replay on the 5 video datasets (with constant $n_x$, not constant memory usage), where network instability is likely to be more problematic due to temporal correlations within video clips and sudden transitions between them during training. 

In both CRUMB and our raw-image replay ablation experiments, the early ``feature extractor'' layers of the CNN are frozen. When we apply CRUMB reconstruction to feature maps from an earlier layer (``early feature replay'' in table~\ref{tab:network_analysis}) and correspondingly allow more layers after this point to have their parameters updated during stream learning, we observe performance worse than both CRUMB and image replay. One interpretation of this is that more unfrozen layers means that more network parameters are exposed to gradient updates and, consequently, to catastrophic forgetting. It is also possible that CRUMB's reconstruction mechanism is best suited to representing abstract, high-level features that are more likely to be found in later CNN layers, and that too much information is lost if CRUMB attempts to represent lower-level information that is interpreted by later layers in more granular ways. 

Given the apparent necessity of preserving the information in feature maps during reconstruction, a surprisingly small codebook of memory blocks is sufficient. As few as 16 memory blocks are needed for optimal performance on CORe50, and CRUMB still performs remarkably well with only 2 memory blocks (see ``number of memory blocks'' experiments in Table~\ref{tab:network_analysis}). Although CRUMB is already highly memory-efficient with the memory blocks themselves occupying negligible space, reducing the number of memory blocks may enable further CPU memory usage optimizations (e.g., 4-bit integers as indices for 16 memory blocks) and also lowers GPU memory usage. Computational and memory efficiency is presumably critical in biological memory systems. Indeed, replay of neuronal activity patterns has been observed to help reinforce and consolidate memories in multiple brain areas across different mammalian species \cite{o2010play,lewis2011overlapping,eichenlaub2020replay}. It is unlikely that neural circuits in the brain use replay mechanisms that preserve as much low-level information as pixel-level replay. Instead, it is interesting to speculate that one of the mechanisms by which brains avoid catastrophic forgetting is by replaying compositions of abstract, high-level features in a manner analogous to CRUMB's replay mechanism.

CRUMB's superior memory and runtime efficiency makes it ideally suited for settings with limited computational resources. Potential applications include edge computing in mobile devices, and autonomous robots that learn continuously from otherwise unmanageable amounts of incoming sensor data while they explore their surroundings. CRUMB could also be used in federated learning contexts, enabling highly effective replay of previously-seen data points via perhaps unrecognizably lossy representations, thereby minimizing both catastrophic forgetting and data security risks. The interpretable qualities of a subset of memory blocks, however, raises the possibility of identifying weak associations with certain generic features contained in a given training example, for a person with unauthorized access to CNN weights, memory blocks, encoded memories of interest, and a reference dataset to discover memory block interpretations. However, it would still be impossible for such a person to completely reconstruct CRUMB's memories in their original encodings (e.g., in pixels). 

CRUMB is implemented here for SqueezeNet \cite{iandola2016squeezenet} and MobileNetV2 \cite{sandler2018mobilenetv2} CNNs, but could be used to mitigate forgetting across different neural network architectures in the future. For example, memory blocks could be used to efficiently reconstruct and replay vector outputs of self-attention heads at intermediate layers in transformer models \cite{vaswani2017attention, dosovitskiy2021image}.

Updating CRUMB's memory blocks using backpropagation in tandem with network weights is highly efficient, and also raises the possibility of tuning memory blocks for shifting domains on the fly. Although updates to the memory blocks beyond pretraining do not appear important for stream learning on CORe50, it is possible that fine-tuning may be necessary in tasks with substantial non-stationarity. Additionally, in this study, CRUMB does not adapt the early \enquote{feature extractor} layers of the CNN during stream learning. However, the early layers could theoretically be trained using the direct prediction loss while the late layers and memory blocks are trained using codebook-out loss or a combination of these two losses: this approach could enable additional flexibility for domain adaptation. Future studies could apply CRUMB to stream learning or reinforcement learning tasks with shifting domains, emulating humans or robots in continuously changing environments. 

\section{Acknowledgments}
\noindent This work was supported by NIH grant R01EY026025, by the National Research Foundation, Singapore under its AI Singapore Programme (AISG Award No: AISG2-RP-2021-025) and its NRFF award NRF-NRFF15-2023-0001, and by the Center for Brains, Minds and Machines, funded by NSF Science and Technology Center Award CCF-1231216. The project was also supported by awards T32GM007753 and T32GM144273 from the National Institute of General Medical Sciences. The content is solely the responsibility of the authors and does not necessarily represent the official views of the National Institute of General Medical Sciences or the National Institutes of Health. 
We would like to thank Trenton Bricken, Spandan Madan, and Warren Sunada-Wong for sharing their helpful insights into our work at various times during this project.

\renewcommand*{\UrlFont}{\rmfamily}
\printbibliography

\vfill

\author{}
\date{}
\title{Supplementary Materials: Tuned Compositional Feature Replays for Efficient Stream Learning}
\maketitle
\beginsupplement

\section{Compositional Replay Using Memory Blocks (CRUMB) outperforms competing algorithms in the class-i.i.d. setting in most cases}

\begin{figure*}[ht]
    \centering
    \begin{minipage}{\textwidth}
        \includegraphics[width=5.1cm, height = 4.0cm]{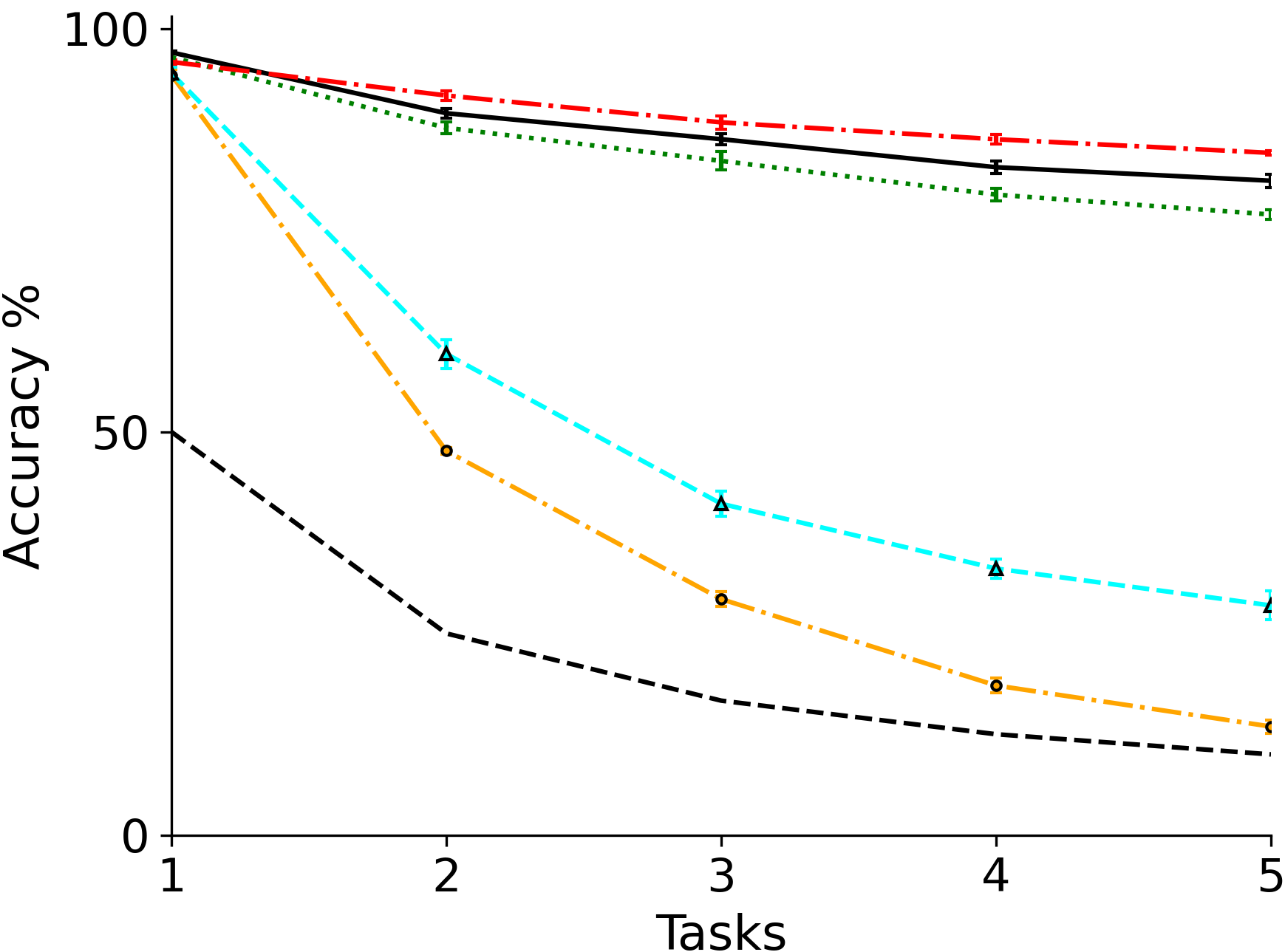}\hfill
        \includegraphics[width=5.3cm, height = 4.0cm]{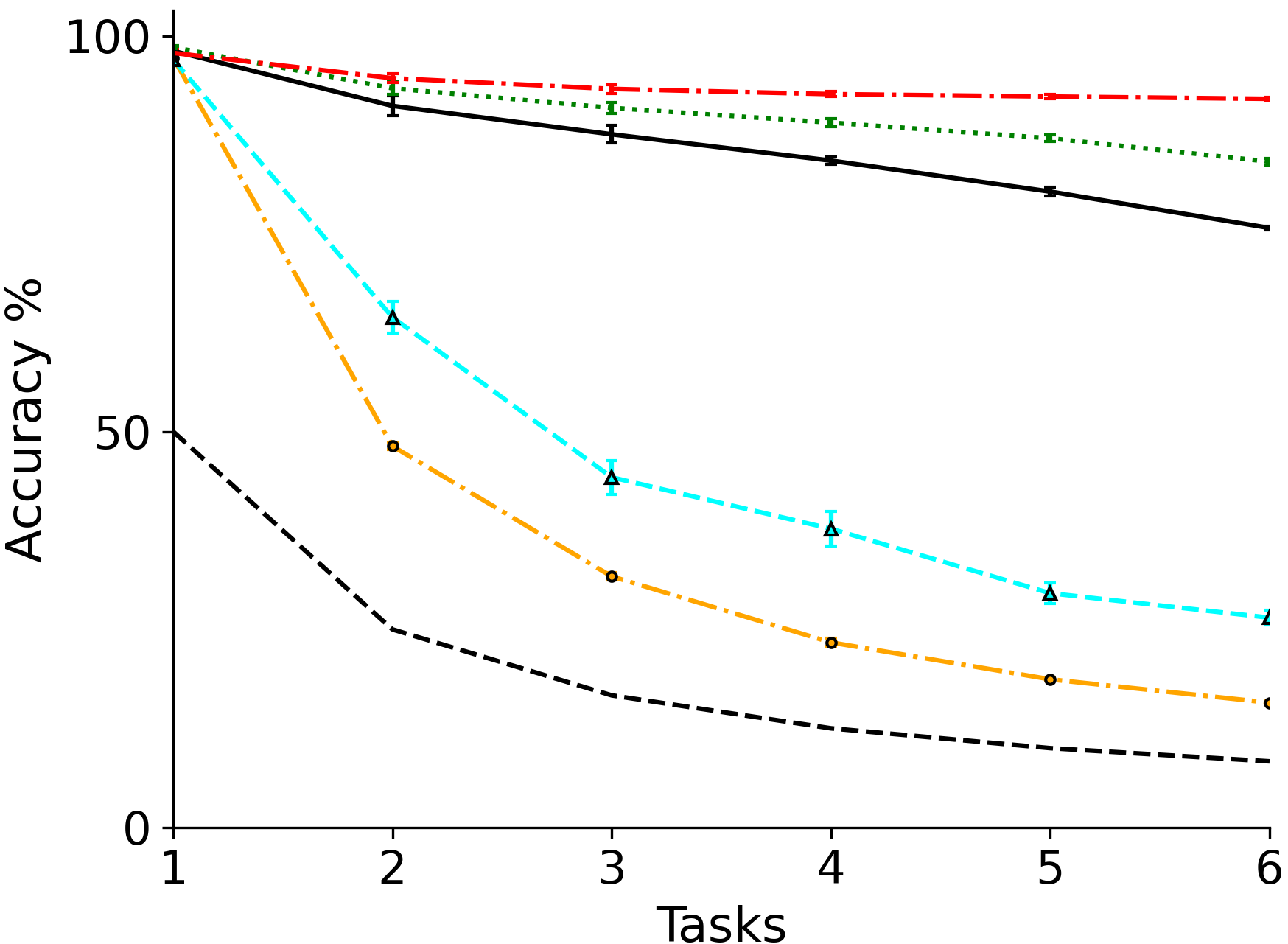} \hfill
        \includegraphics[width=5.6cm, height = 4.0cm, trim = 10mm 1.5mm 0 0]{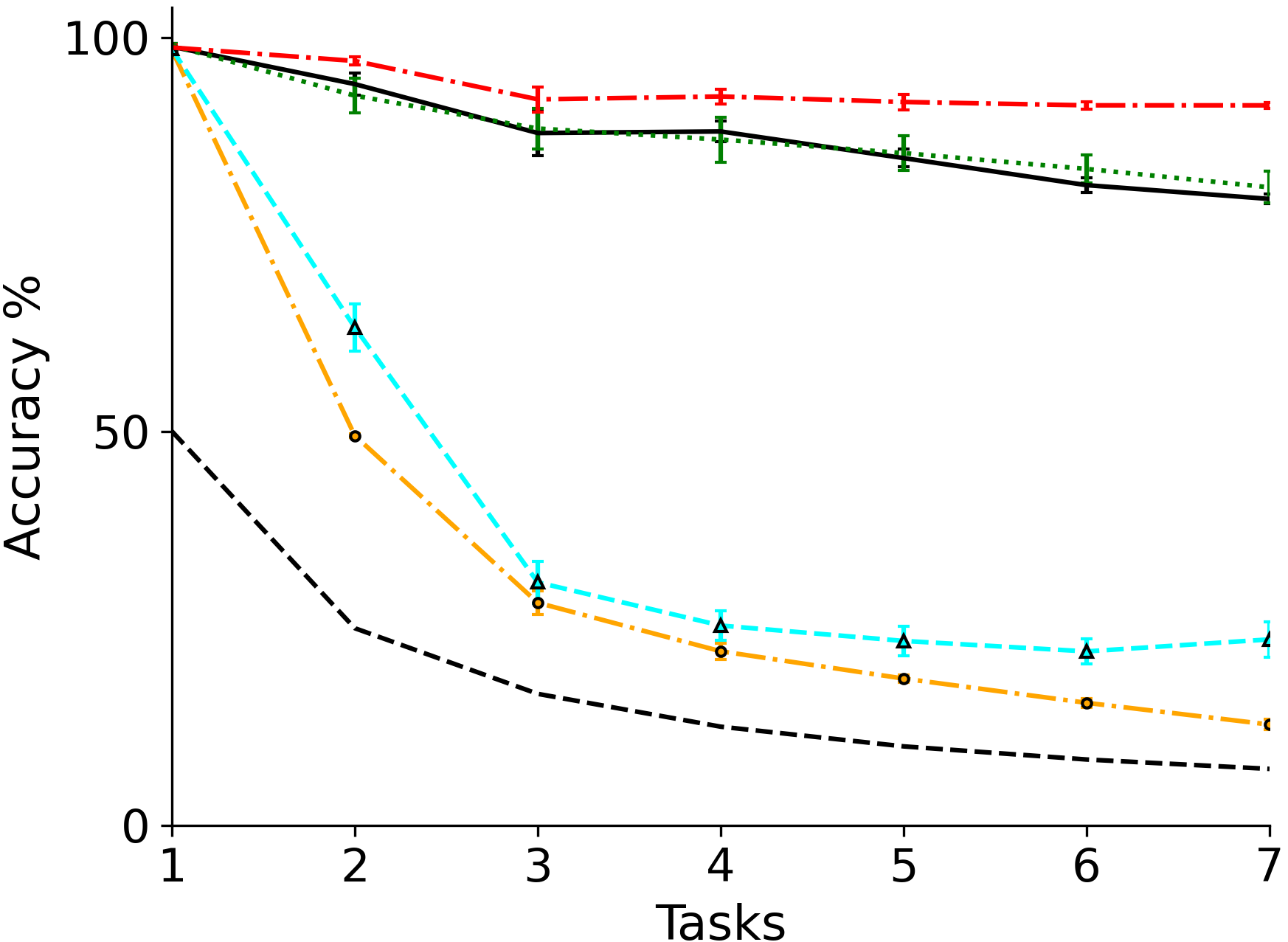}  \hfill
        
    \hspace{1.5cm} a. CORe50 (class-i.i.d.) \hspace{2.2cm} b. Toybox (class-i.i.d.) \hspace{3.2cm} c. iLab  (class-i.i.d.)
    \\
    \end{minipage}\hfill

    \vspace{-2mm}

    \begin{minipage}{\textwidth}
        \includegraphics[width=6.4cm, height = 4.3cm]{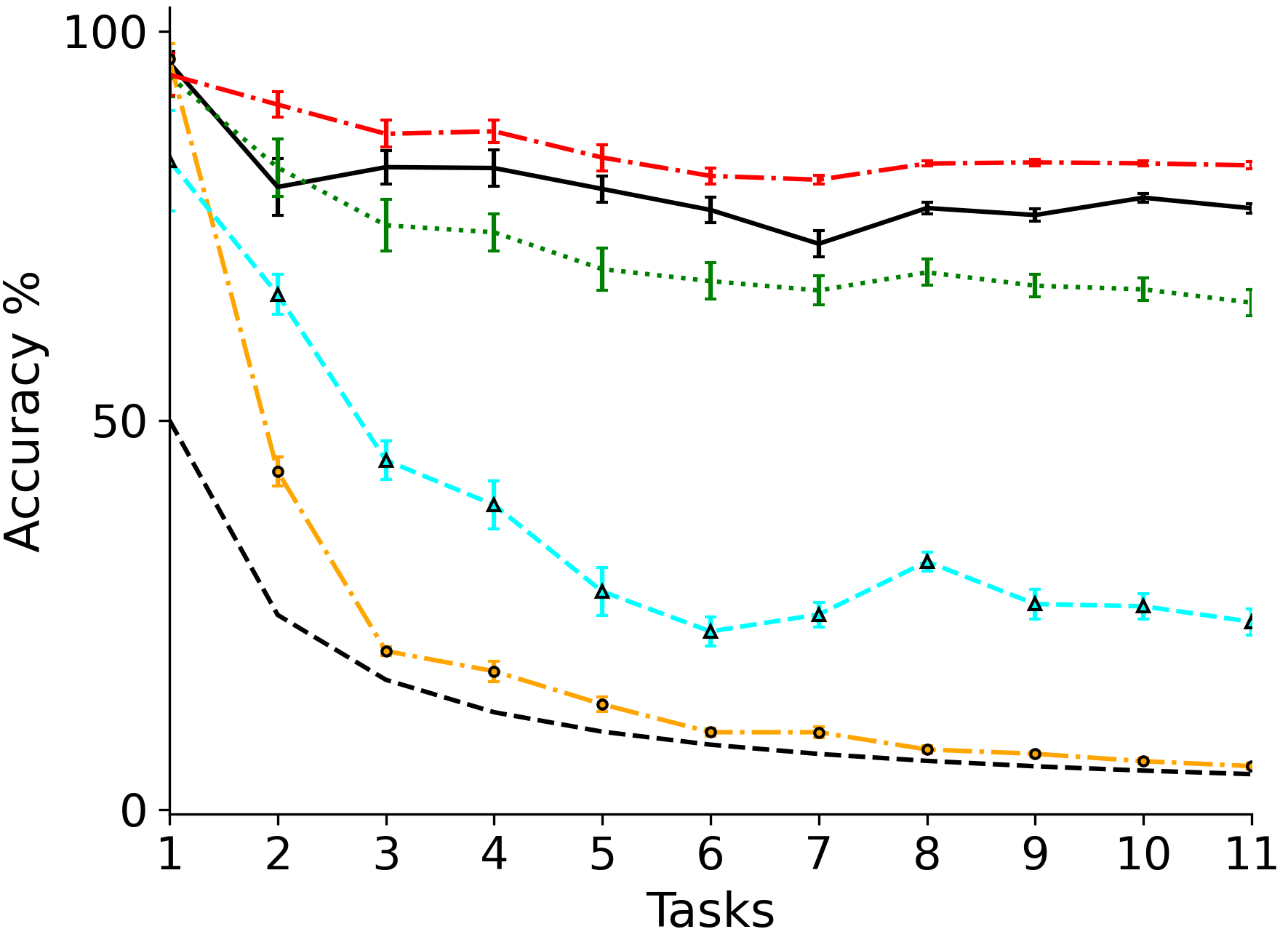}\hfill
        \includegraphics[width=6.4cm, height = 4.3cm]{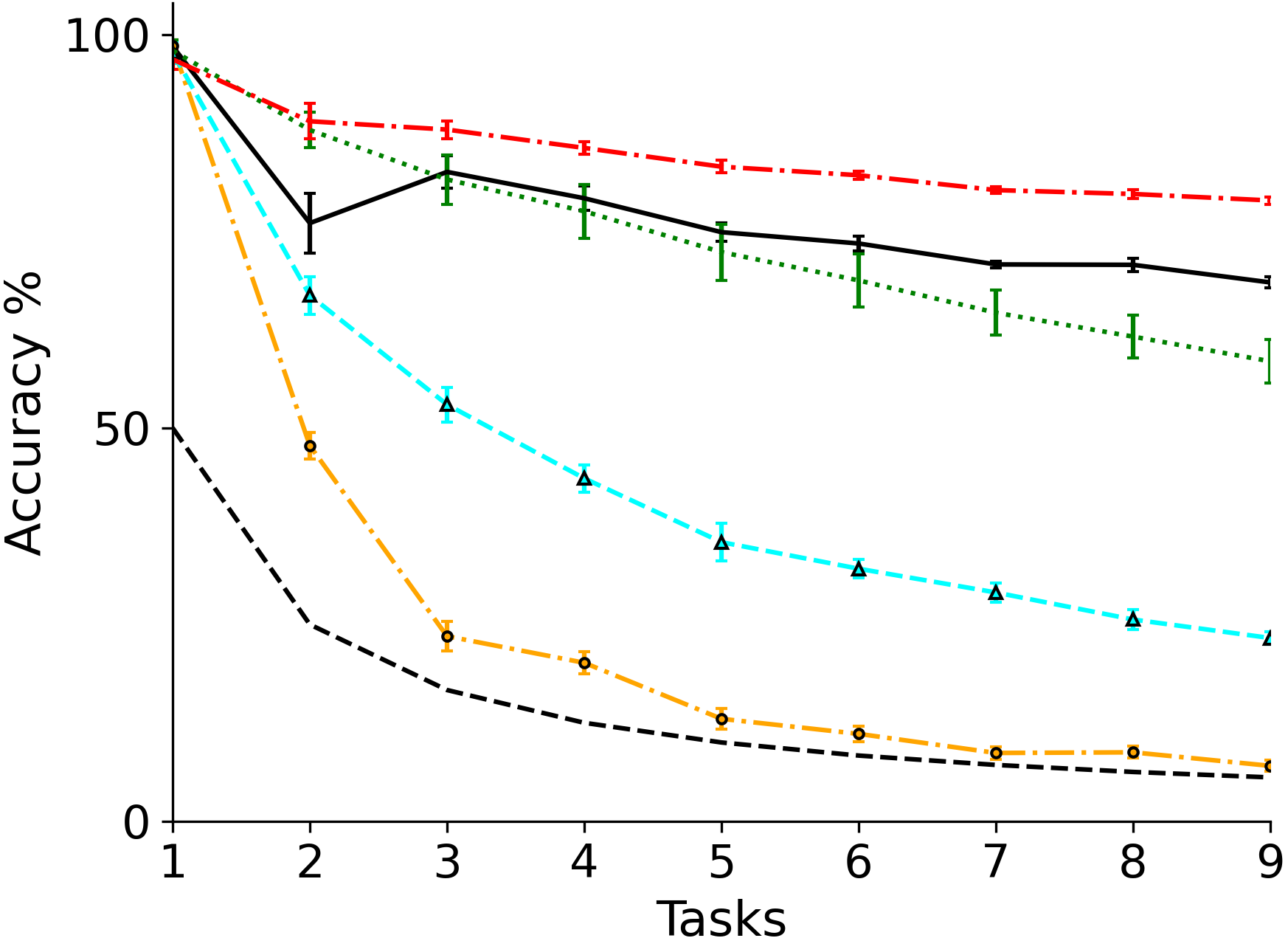} \hfill
        \includegraphics[height = 3.0cm]{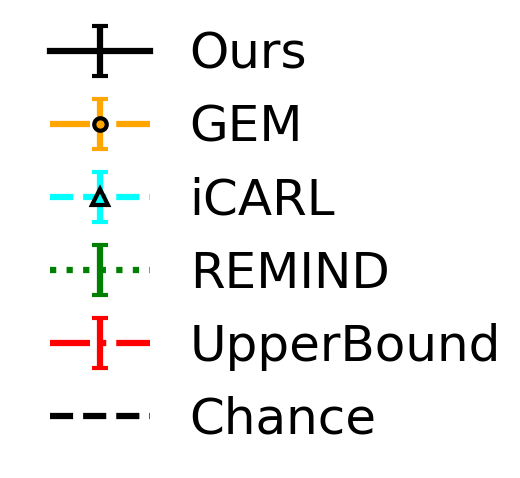}\hfill
        
    \hspace{1.2cm} d. iLab + CORe50 (class-i.i.d.)  \hspace{3.0cm} e. iCub  (class-i.i.d.) \hspace{3.3cm} f. Legend  
    \\
    \end{minipage}\hfill

    \caption{\textbf{In the class-i.i.d. setting, CRUMB outperforms most baseline algorithms and performs near the the upper bound on some datasets}. Line plots show top-1 accuracy in online stream learning on video datasets (a) CORe50, (b) Toybox, (c) iLab, (d) iLab + CORe50, and (e) iCub in the class-i.i.d. setting. All models train on the first task for many epochs, but view each image only once on all subsequent tasks. Accuracy estimates are the mean from 10 runs, where each run has different class and image/video clip orderings.
    Error bars show the root-mean-square error (RMSE) among runs. Results for all baselines are in Table~\ref{tab:all_results} in the main text.}
\label{fig:video_streaming_class_i.i.d.}
\end{figure*}

We report top-1 accuracy results (measured on all tasks/classes in each dataset at the end of stream learning training) for CRUMB and all competing baseline algorithms on five video streaming datasets (CORe50 \cite{lomonaco2017core50}, Toybox \cite{wang2018toybox}, iLab \cite{borji2016ilab}, iLab+CORe50, and iCub \cite{pasquale2016object}) and two image datasets (Online-CIFAR100 \cite{krizhevsky2009learning}, Online-Imagenet \cite{deng2009imagenet}) in both class-i.i.d. and class-instance training protocols in Table~\ref{tab:all_results} in the main text. For CRUMB and a subset of baseline algorithms, we illustrate task-by-task top-1 accuracy (on all previously seen classes) for the five video datasets in the class-i.i.d. 
setting in Fig.~\ref{fig:video_streaming_class_i.i.d.}. Class-instance plots for video datasets, and class-i.i.d. plots for image datasets, are in main text Fig.~\ref{fig:video_streaming_main}. 

\section{CRUMB's performance is competitive with baseline algorithms even when memory buffer size is unlimited} \label{sec:unlimited-memory_crumb_vs_baselines}

\begin{figure*}[ht]
    \centering
    \begin{minipage}{\textwidth}
        \includegraphics[height=5.0cm]{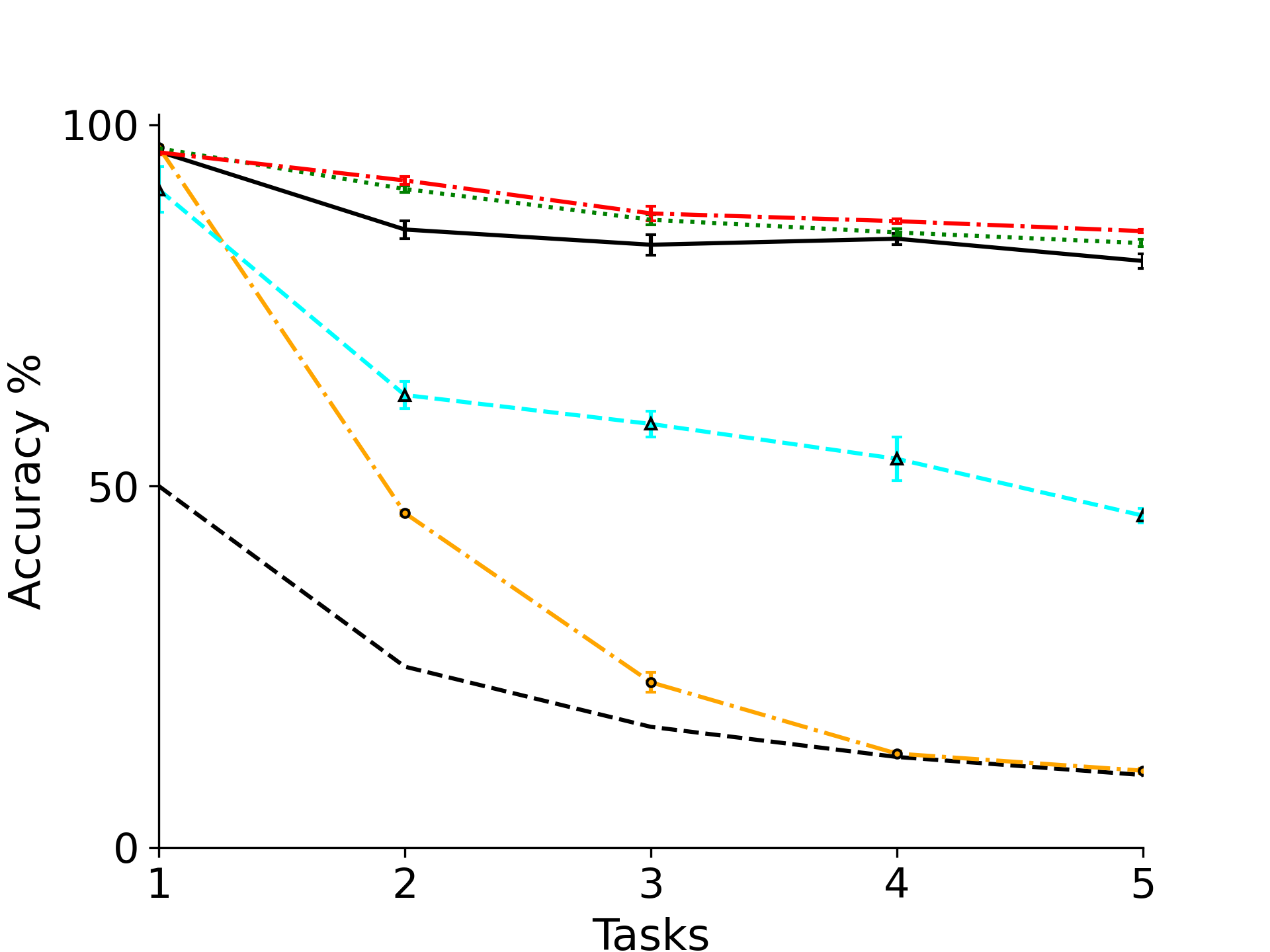}\hfill
        \includegraphics[height = 5.0cm]{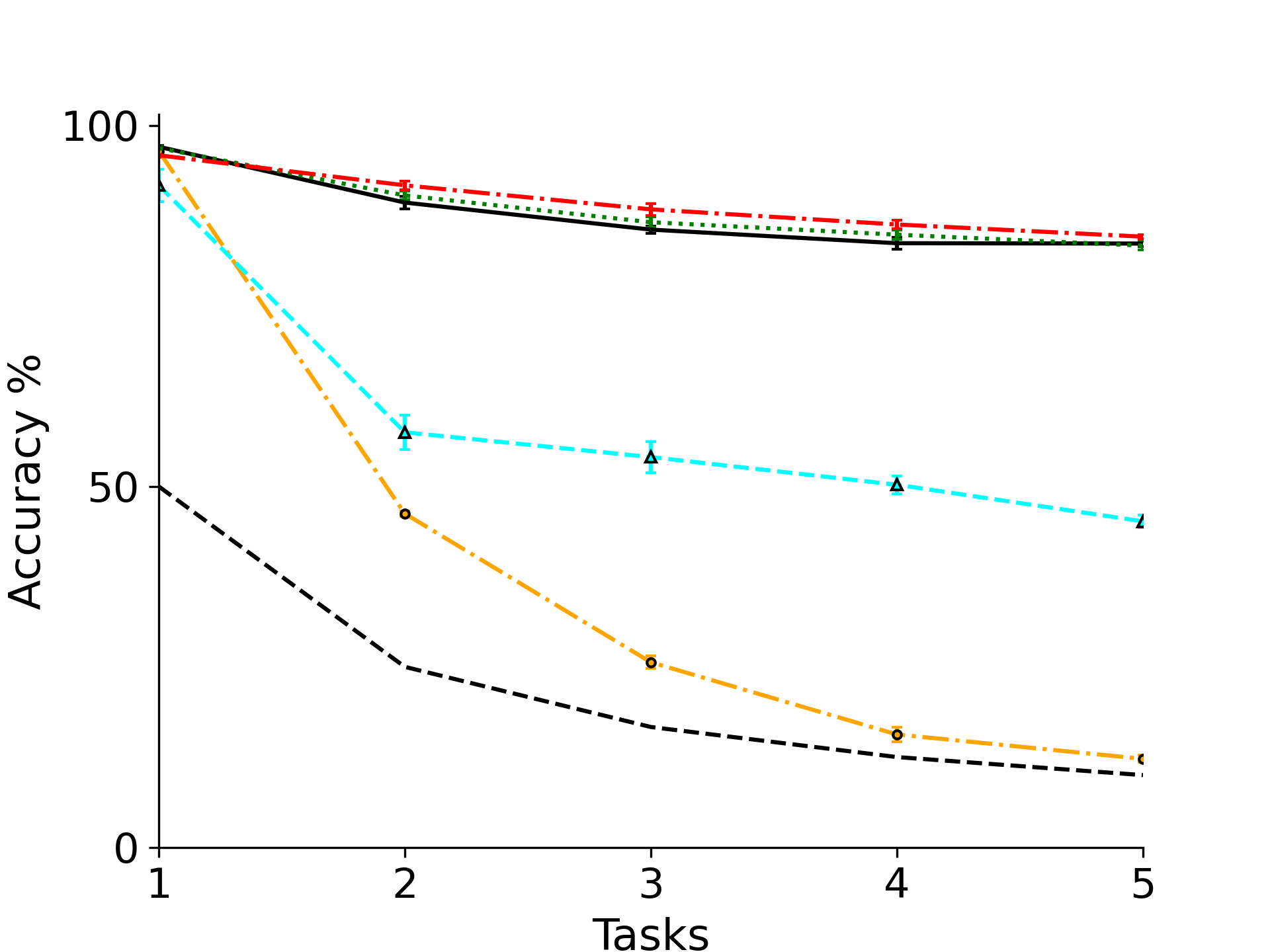} \hfill
        \includegraphics[height = 3.0cm]{Supp_figures/Legend.png}  \hfill
        
    \hspace{1.3cm} a. CORe50 (class-instance) \hspace{3.4cm} b. CORe50 (class-i.i.d.) \hspace{3.0cm} c. Legend
    \end{minipage}\hfill

    \caption{\textbf{CRUMB attains competitive levels of performance in conditions of unlimited memory usage}. Line plots show top-1 accuracy in online stream learning on the video dataset CORe50. For this comparison among replay methods, all models are allowed to store all previously encountered images in a replay buffer and intersperse them with images encountered while training on new tasks. All models train on the first task for many epochs, but view each image only once on all subsequent tasks. Accuracy estimates are the mean from 10 runs, where each run has different class and image/video clip orderings. Error bars show the root-mean-square error (RMSE) among runs.}
\label{fig:video_streaming_unlimited_memory}
\end{figure*}

Our primary baseline comparison experiments in the main text focus on comparing CRUMB with competing algorithms under a fixed memory budget: methods that store and replay entire raw images cannot store as many training examples, affecting their continual learning performance. CRUMB is specifically designed with memory-constrained conditions in mind, and compresses each training example stored in its replay buffer to occupy only 3.6\% as much memory as an entire image (e.g., as stored by image-based replay baseline iCARL \cite{rebuffi2017icarl}). Nonetheless, we demonstrate here in Fig.~\ref{fig:video_streaming_unlimited_memory} that CRUMB obtains competitive performance even when memory usage is unlimited, greatly outperforming iCARL. CRUMB's accuracy after training on all tasks is slightly lower than that of REMIND \cite{hayes2020remind} in the class-instance setting under these conditions, with both methods close to the offline upper bound in both class-instance and class-i.i.d.

\section{CRUMB maintains its bias towards object shape information after class-i.i.d. stream learning on video datasets} \label{sec:class_i.i.d._shape_bias}

\begin{figure}[t]
    \centering
    \includegraphics[width=\linewidth]{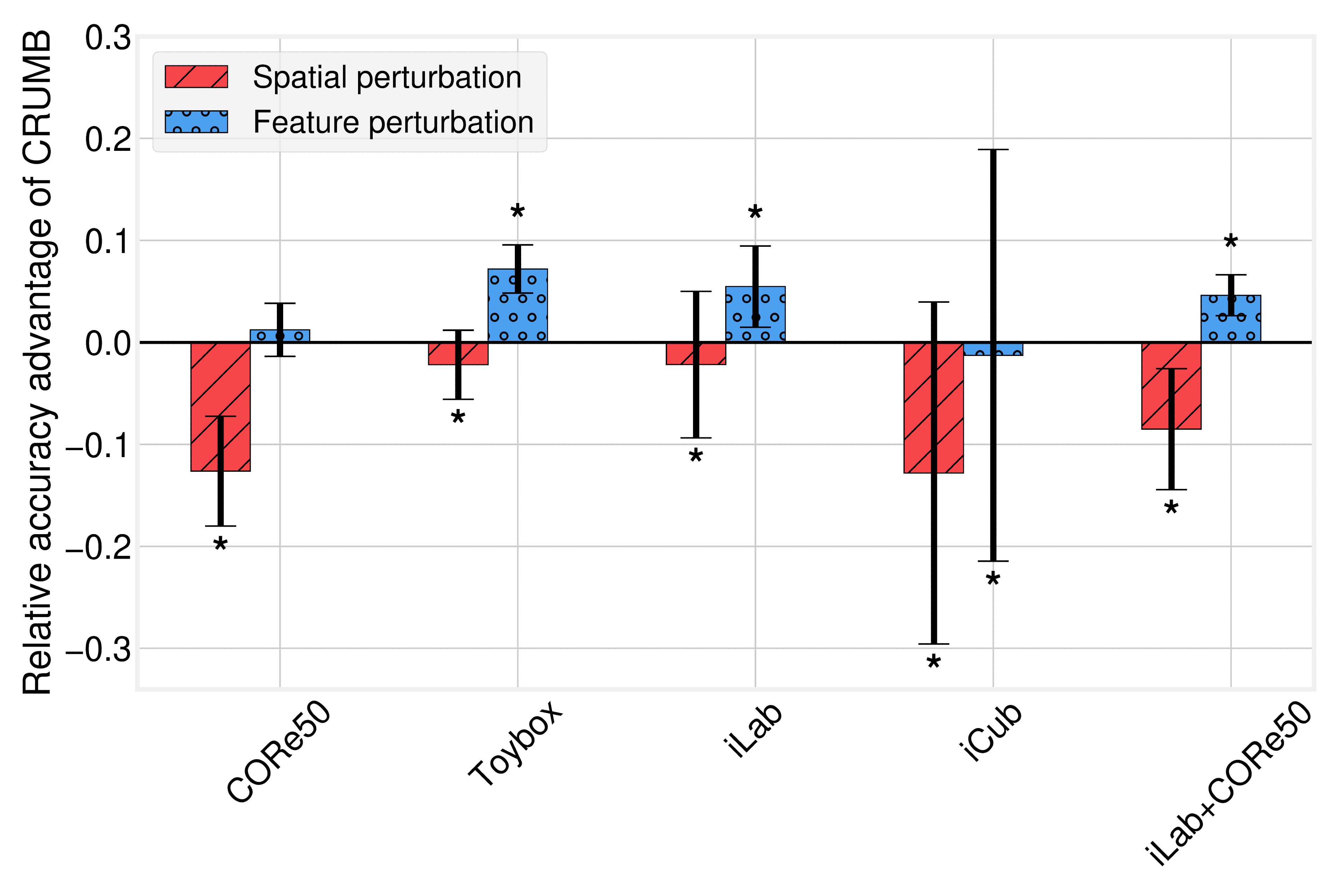}
    \vspace{-8mm}
    \caption{\textbf{The bias towards shape information induced by CRUMB pretraining persists through stream learning in the class-i.i.d. setting}. The height of each bar shows how much smaller (or larger, if negative) CRUMB's drop in normalized test set accuracy under a perturbation is, in comparison to a control network. ``Relative accuracy advantage'' is calculated by dividing the difference in accuracy caused by a perturbation by the unperturbed accuracy, and then subtracting this result for CRUMB from that of the control network (see main-text Section~\ref{sec:shape_bias}).
    ``Spatial perturbation'' shuffles the spatial positions of all feature vectors in an intermediate feature map (at the same layer where it is reconstructed by CRUMB), ``feature perturbation'' randomly sets half of the feature map's features to zero, and ''style perturbation'' uses images from Stylized-ImageNet \cite{geirhos2018}. Streaming results (to the right of grey dotted line) are in the class-i.i.d. setting: class-instance shape-texture bias results for the video datasets, and class-i.i.d. results for Online-CIFAR100 and Online-ImageNet, are available in Fig.~\ref{fig:shape_bias_class_instance} in the main text. Error bars are standard errors of the mean of relative accuracy advantage among or 10 independent runs. * denotes a statistically significant difference from 0 (see Section~\ref{sec:data_analysis}).
    }
	\label{fig:shape_bias_class_i.i.d.}
\end{figure}

In Section~\ref{sec:shape_bias} and Fig.~\ref{fig:shape_bias_class_instance} in the main text, we observe that pretraining CRUMB on ImageNet \cite{deng2009imagenet} induces a bias in the CNN towards attending to object shape information more than image texture information, an effect that has been shown to mitigate catastrophic forgetting by flattening the loss minimum of each task \cite{shi2022robustness}. Fig.~\ref{fig:shape_bias_class_instance} in the main text visualizes the extent of this ``shape bias'' for CRUMB trained on video datasets in the class-instance setting. Fig.~\ref{fig:shape_bias_class_i.i.d.} shows results in the class-i.i.d. setting. As for class-instance, we observe that CRUMB mostly retains its shape bias during class-i.i.d. stream learning. 

\section{Data analysis} \label{sec:data_analysis}

\subsection{Data cleaning}
For our main results on the video datasets CORe50, Toybox, and iLab, we noticed that a small subset of runs for some models had markedly reduced accuracy on the first task compared to other runs. To facilitate fair comparisons among models, we excluded all runs with an initial task accuracy less than 80\% from all analysis and results. For the small number of algorithm/dataset/protocol combinations for which no runs exceeded 80\% on the first task, we filtered at a 60\% threshold, or a 40\% threshold if no runs exceeded 60\%. We did not encounter this issue for any runs of CRUMB on any dataset, or for any method on Online-CIFAR100 and Online-Imagenet.

\subsection{Statistics for model analysis experiments}
Our model analysis experiments in main-text Section~\ref{sec:model_analysis} compared the performance of CRUMB with various ablated or otherwise perturbed versions of CRUMB. For each comparison with the original algorithm, we evaluated statistical significance of pairwise differences using the following method:
\begin{enumerate}[i.]
\item Divide the test set from the dataset being used into batches of 100 images. The images should be randomly sampled without replacement, and the sampling should be done only once (or, using a fixed random seed) for all experiments such that each version of the algorithm is evaluated on the exact same batches of images.
\item Evaluate CRUMB and each experimentally perturbed version of CRUMB on the same set of image batches and record mean top-1 accuracy on each batch. This is done for each of the 5 independent training runs, and accuracies are pooled across runs. Therefore, for each training protocol (class-instance and class-i.i.d., for which all analyses are kept separate), each version of the algorithm has $n_r \times n_b$ top-1 accuracy estimates, where $n_r$ is the number of runs and $n_b$ is the number of 100-image batches in the test set. Conceptually, we treat the accuracy on each batch as an independent sample indicating the accuracy of the corresponding algorithm on a roughly continuous scale, with each run of each algorithm tested on the exact same batches of images. 
\item Perform a paired-samples t-test for each comparison, using accuracy on each image batch of CRUMB and the perturbed version of CRUMB as a sample pair and pooling sample pairs across runs. We used a global p-value cutoff of $p < 0.01$ to report the statistical significance of t-test results for each comparison between CRUMB and a perturbed version of CRUMB. 
\end{enumerate}

For our experiments on shape-texture bias, we employ a similar approach by first calculating accuracy on batches of 100 images at a time. For each batch, we subtract the model's perturbed accuracy (i.e., after spatial, feature, or style perturbation, see Section~\ref{sec:shape_bias} in the main text) from the unperturbed accuracy, and divide the result by the unperturbed accuracy to obtain the relative accuracy drop for each perturbation. We compare the relative accuracy drops for CRUMB and a control network on all batches, pooling across runs with different data orderings, using the Wilcoxon signed-rank test for paired samples. We apply a global p-value cutoff of $p < 0.01$ to report the significance of any differences, visualized as CRUMB's relative accuracy advantage being either above or below zero in main-text Fig.~\ref{fig:shape_bias_class_instance} and supplementary Fig.~\ref{fig:shape_bias_class_i.i.d.}.

\section{Replay buffer size calculations} \label{sec:replay_buffer_size_calc}

For replay-based baseline algorithms, we limit the number of examples that can be stored in the buffer to fit within a memory budget that is held constant for all methods in our main results (main text Table~\ref{tab:all_results}). We do not apply this constraint for weight regularization approaches. To calculate the maximum number of training examples we can store in the replay buffer for each experiment, we first set the number of examples $n_{\text{raw}}$ that raw-image replay methods such as iCARL may store, then calculate how many examples ($n_x$) CRUMB can fit into the same amount of memory using the formula: 

\begin{equation}
n_x = \frac{n_r (3 w_i h_i) - bd}{swh/d}
\end{equation}

Where $w_i$ and $h_i$ are raw image width and height respectively ($224 \times 224$ for our experiments), the codebook matrix has dimensions $b \times d$ ($b$ memory blocks, each of dimension $d$), and the feature map has dimensions $s \times w \times h$ ($s$ features in a $w \times h$ spatial grid). The numerator corresponds to the number of 8-bit RGB values needed to store one image, subtracting 
a discounting factor for the number of values in the memory blocks themselves.
The denominator corresponds to the number of 8-bit integer indices required to encode one feature map. Concretely, the memory budgets are 2.2 MB on CORe50, Toybox, and iLab, 14.3 MB on CIFAR100, and 1.44 GB on ImageNet based on the number of 8-bit integers each method stores per training example. 

For direct comparisons between algorithms in our main results, we applied both CRUMB and REMIND to the SqueezeNet network architecture \cite{iandola2016squeezenet}. To calculate $n_x$ for REMIND, we multiplied the compression ratio provided by the REMIND paper (959,665 feature maps/10,000 raw images) by the ratio of values in one feature map from ResNet18 (used in the REMIND paper, $512 \times 7 \times 7$) to those in one feature map from SqueezeNet ($512 \times 13 \times 13$)  \cite{hayes2020remind}. We then multiplied the resulting ratio of 278,246 feature maps/10,000 raw images by $n_{\text{raw}}$ to obtain the corresponding $n_x$ for each dataset.


\end{document}